\documentclass[final,3p,times,twocolumn]{elsarticle}

\usepackage{lineno}
\usepackage{hyperref}
\usepackage{framed,multirow}
\usepackage{amssymb}
\usepackage{amsmath}
\usepackage{latexsym}
\usepackage{url}
\usepackage{xcolor}
\usepackage{glossaries}
\usepackage{cite}
\usepackage{array}
\usepackage{graphicx,caption,rotating}
\usepackage{setspace}
\usepackage{lscape}
\usepackage{soul}
\usepackage{subcaption}
\usepackage[labelformat=parens,labelsep=quad,skip=3pt]{caption}
\usepackage{graphicx}
\modulolinenumbers[5]

\definecolor{nblue}{rgb}{0,0,0}
\definecolor{nred}{rgb}{0,0,0}

\newacronym{3dcnn}{3D CNN}{3D Convolutional Neural Network}
\newacronym{amt}{AMT}{Amazon Mechanical Turk}
\newacronym{bof}{BoF}{Bag-of-Features}
\newacronym{c3d}{C3D}{Convolutional 3D Network}
\newacronym{cca}{CCA}{Canonical Correlation Analysis}
\newacronym{ccv}{CCV}{Columbia Consumer Videos}
\newacronym{cnn}{CNN}{Convolutional Neural Network}
\newacronym{conssev}{ConSSEV}{Convex Combination of Similar Semantic Embedding Vectors}
\newacronym{dap}{DAP}{Direct Attribute Prediction}
\newacronym{dda}{DDA}{Data-driven attributes}
\newacronym{dtf}{DTF}{Dense Trajectory Features}
\newacronym{ecoc}{ECOC}{Error Correcting Output Codes}
\newacronym{fps}{fps}{frames per second}
\newacronym{fsl}{FSL}{Few-Shot Learning}
\newacronym{fv}{FV}{Fisher Vectors}
\newacronym{gan}{GAN}{Generative Adversarial Network}
\newacronym{gcn}{GCN}{Graph Convolutional Network}
\newacronym{gmil}{GMIL}{Generalised Multiple Instance Learning}
\newacronym{gru}{GRU}{Gatet Recurrent Unit}
\newacronym{hog}{HoG}{Histogram of Gradient}
\newacronym{hof}{HoF}{Histogram of Optical Flow}
\newacronym{har}{HAR}{Human Action Recognition}
\newacronym{i3d}{I3D}{Inflated 3D Network}
\newacronym{iaf}{IAF}{Inverse Autoregressive Flow}
\newacronym{iap}{IAP}{Indirect Attribute Prediction}
\newacronym{idt}{IDT}{Improved Dense Trajectories}
\newacronym{itf}{ITF}{Improved Trajectory Features}
\newacronym{jsd}{JSD}{Jansen-Shannon Divergence}
\newacronym{kliep}{KLIEP}{Kullback-Leibler Importance Estimation Procedure}
\newacronym{lda}{LDA}{Latent Dirichlet Allocation}
\newacronym{lr}{LR}{Least Square Regression}
\newacronym{lstm}{LSTM}{Long Short-Term Memory}
\newacronym{m2latm}{M2LATM}{Multi-Modal Latent Attribute Topic Model}
\newacronym{map}{MAP}{Maximum-a-Posteriori Estimation}
\newacronym{mbh}{MBH}{Motion Boundary Histogram}
\newacronym{mle}{MLE}{Maximum Likelihood Estimation}
\newacronym{mlp}{MLP}{Multi-Layer Perceptron}
\newacronym{mse}{MSE}{Mean Squared Error}
\newacronym{mtl}{MTL}{Multi-Task Learning}
\newacronym{nmf}{NMF}{Non-Negative Matrix Factorization}
\newacronym{pca}{PCA}{Principal Component Analysis}
\newacronym{ptm}{PTM}{Probabilistic Topic Model}
\newacronym{svm}{SVM}{Support Vector Machine}
\newacronym{svr}{SVR}{Support Vector Regression}
\newacronym{ucf}{UCF}{University of Central Florida}
\newacronym{ur}{UR}{Universal Representation}
\newacronym{zsar}{ZSAR}{Zero-Shot Action Recognition}
\newacronym{zsl}{ZSL}{Zero-Shot Learning}

\DeclareMathOperator*{\argmax}{arg\,max}

\makeatletter
\def\ps@pprintTitle{%
 \let\@oddhead\@empty
 \let\@evenhead\@empty
 \def\@oddfoot{}%
 \let\@evenfoot\@oddfoot}
\begin{document}
\tolerance=999
\sloppy

\begin{frontmatter}
\title{Zero-Shot Action Recognition in Videos: A Survey}
\author[1,3]{Valter Estevam\corref{mycorrespondingauthor}}
\cortext[mycorrespondingauthor]{Corresponding author}
\ead{valter.junior@ifpr.edu.br}
\author[2]{Helio Pedrini}
\author[3]{David Menotti}
\address[1]{Federal Institute of Paran\'a, Irati-PR, 84500-000, Brazil}
\address[2]{Universiy of Campinas, Institute of Computing, Campinas-SP, 13083-852, Brazil}
\address[3]{Federal University of Paran\'a, Department of Informatics, Curitiba-PR, 81531-970, Brazil}

\begin{abstract}
Zero-Shot Action Recognition has attracted attention in the last years and many approaches have been proposed for recognition of objects, events and actions in images and videos. There is a demand for methods that can classify instances from classes that are not present in the training of models, especially in the complex problem of automatic video understanding, since collecting, annotating and labeling videos are difficult and laborious tasks. We have identified that there are many methods available in the literature, however, it is difficult to categorize which techniques can be considered state of the art. Despite the existence of some surveys about zero-shot action recognition in still images and experimental protocol, there is no work focused on videos. Therefore, we present a survey of the methods that comprise techniques to perform visual feature extraction and semantic feature extraction as well to learn the mapping between these features considering specifically zero-shot action recognition in videos. We also provide a complete description of datasets, experiments and protocols, presenting open issues and directions for future work, essential for the development of the computer vision research field.
\end{abstract}
\end{frontmatter}
%\linenumbers

\section{Introduction}
\label{sec:introduction}

In recent years, many works in the computer vision field have explored the human action or activity recognition problem using still images or videos. Some authors, such as~\citet{turaga:2008-ref61}, define actions as simple motion patterns often performed by only one human being, and activities as more complex patterns that involve coordinated actions of a small group of humans. However, there is no universal understanding of these concepts in the literature. 
\textcolor{nblue}{In this text, we adopt the term action recognition to refer to both concepts, regardless of whether the authors consider their work as action or activity recognition. Following this assumption, several surveys~\citep{turaga:2008-ref61, poppe:2010-ref62, aggarwal:2011-ref63, guo:2014-ref64, ziaeefard:2015-ref10, kongfu:2018-ref29} show approaches addressing the \gls{har} problem by proposing new visual or semantic features describing the actions more accurately}. 
For example, the \gls{dtf}~\citep{wang:2013a} and its variant, the \gls{idt}~\citep{wang:2016}, are two  successful methods based on handcrafted visual features. 
Another group of works explores semantic features, such as poses and poselets~\citep{agahian:2020}, objects~\citep{nazli:2010}, scenes~\citep{zhang:2014} and attributes~\citep{zhang:2013}, or investigates new inference methods, such as in~\citep{liushu:2018}. 
Recently, deep learning has been applied to \gls{har}, leveraging visual features through the exploration of convolution operation, temporal modeling, and multi-stream configuration, as shown by~\citet{kongfu:2018-ref29}.

All these approaches suffer from inherent drawbacks, for example: (i) they do not generalize very well on large and complex datasets, such as Charades~\citep{sigurdsson:2016} or Kinetics~\citep{carreira:2017}; (ii) the handcraft visual features are very expensive to compute; (iii) manual-annotated semantic features require heavy human labor or expert knowledge, which are not always available; and (iv) many labeled examples are required to reduce the generalization problem when deep learning is used.

\begin{figure*}[!htb]
	\centering
	\includegraphics[width=0.98\textwidth]{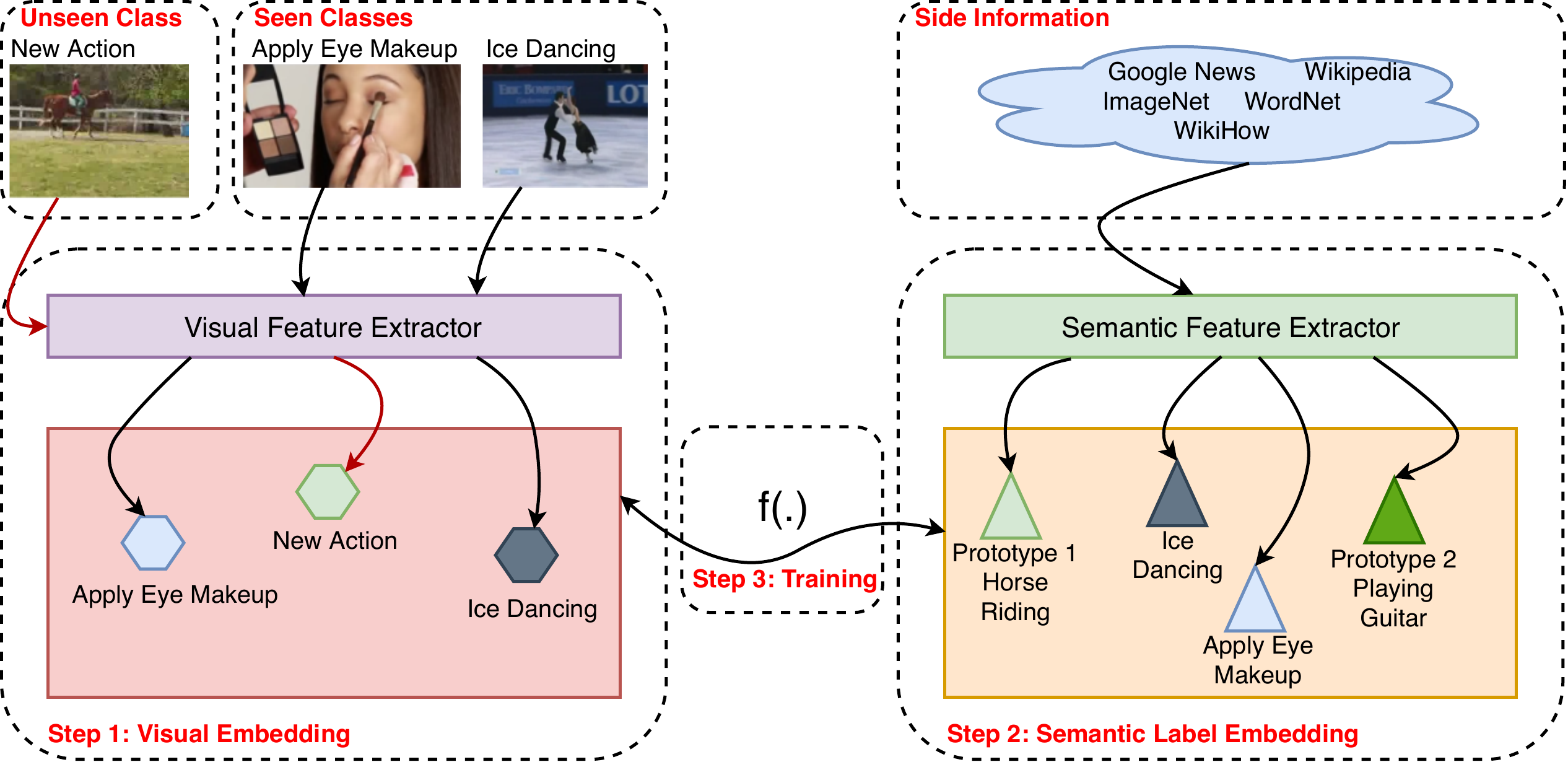}
	\caption{Schematic representation of a ZSL human action recognition framework.}
	\label{fig:esquemageral}
\end{figure*}

In a real-world scenario, there are many more actions than in the academic benchmark datasets used to learn the models. Moreover, the new examples may be unlabeled, which makes the supervised methods inappropriate. In this context, \gls{zsl} emerges attempting to overcome these limitations. 

The human ability to recognize an action without ever having seen it before, that is, associating semantic information from several sources to the visual appearance of actions, is the inspiration of \gls{zsl} approaches~\citep{kodirov:2015-ref22}. \textcolor{nblue}{In Figure~\ref{fig:esquemageral}, we provide an overview of \gls{zsl} approaches considering the application in videos. This general scheme can also be found in \gls{zsl} applied to object and event recognition in both images and videos~\citep{fu:2018-ref30}. We introduce the main aspects of the approaches throughout this text}.

In this example, some videos from \texttt{Apply Eye Makeup} and \texttt{Ice Dancing} action classes are used to extract visual features in order to compose a visual space. Commonly, these visual features are obtained using the \gls{idt} method~\citep{wang:2013a}, \gls{hog}, \gls{hof}, and \gls{mbh} algorithms with Bag-of-Features approach~\citep{rohrbach:2013-ref39}; or using deep features from \gls{c3d}~\citep{tran:2015} or \gls{i3d}~\citep{carreira:2017}.

In \gls{zsl}, we assume that we have a set of all possible action class labels, and, for some of them, there is no video example. Therefore, auxiliary semantic side information is required to provide a computational representation for the labels. This representation usually relies on attributes manually annotated~\citep{qiu:2011-ref11}, word vectors~\citep{kodirov:2015-ref22} and hierarchical structures~\citep{alnasser:2018-ref44}, which are called prototypes. If we try to recognize a new video from the \texttt{New Action} class, which has never been seen before, in addition to extracting visual features, it is necessary to associate them with a suitable prototype and assign a label. This is made by learning an $f(\cdot)$ \textit{mapping function} between these spaces. As discussed in Section~\ref{sec:trainingstep}, this mapping function can assume several ways to be performed directly into the semantic space, indirectly by creating an intermediate space or directly into the visual space.

\textcolor{nblue}{Thus, we concentrate our investigation in approaches that address the problem of recognizing human actions, without having seen them before, in small video clips, typically with less than 10 seconds. This is referred to as \gls{zsar} problem\footnote{\textcolor{nblue}{Throughout this text, when the term ZSAR is used, it refers to ZSAR in videos, rather than ZSAR in images. The latter is not widely studied and its approaches are more similar to ZSL applied to object recognition than the ZSAR techniques covered by this text.}}. We do not take into account the \gls{fsl} task since it is a different problem. In \gls{fsl}, the presence of some examples usually introduces a significant disturbance in the probability distribution of the representations, which degrades the performance over both class groups (i.e., with many and few examples). Works focused on \gls{fsl} usually perform a matching between a query video and the representative videos of each class, and the general problem is how to create better representations in order to allow this matching. Some examples are shown in~\citep{bishay:2019-ref78,ghosh:2020-ref74,zhu:2018-ref55}}.

There exist other surveys related to \gls{zsl}. For instance, \citet{fu:2018-ref30} and~\citet{xian:2017-ref17} provided an overview of \gls{zsl} problems, especially about still images and experimental protocols. More recently, \citet{wang:2019} investigated the \gls{zsl} paradigm with focus on settings, methods and applications for actions, objects and events. 
\textcolor{nblue}{Although some initial works in \gls{zsar} adopted approaches inspired by zero-shot object recognition, covered by other surveys, there are several approaches specially designed for \gls{zsar} that deserve attention}.
To the best of our knowledge, there is no survey focused on \gls{zsar} in videos and our main contributions are three-fold: (i) to provide a complete description of \gls{zsl} methods applied to human action recognition in videos detailing the methods used to extract visual features, semantic features, as well to perform the training; (ii) to present a discussion about the limitations of the benchmark datasets and evaluation protocols adopted in works in the literature; and (iii) to identify open issues pointing future research strategies, based on the natural evolution of the ZSAR approaches, and inspire different approaches in this knowledge domain.

The remainder of the text is organized as follows. We review the methods used to perform visual and semantic embedding in Section~\ref{sec:visualsemanticembedding} and provide a complete description of \gls{zsar} approaches in Section~\ref{sec:trainingstep}. The benchmark datasets are presented in Section~\ref{sec:datasetsprotocol}, whereas experimental protocols and performance are discussed in Section~\ref{sec:performance}. We discuss open issues and directions for future work in Section~\ref{sec:open}. Finally, some concluding remarks are presented in Section~\ref{sec:conclusion}.

\section{Visual and Semantic Label Embedding Steps}
\label{sec:visualsemanticembedding}

Two crucial steps in any \gls{zsar} method are the visual and semantic label embeddings. They are responsible for providing the features used to map the visual appearance to the semantic description of actions. 

\subsection{Visual Embedding Step}
\label{sec:visualembedding}

\begin{figure*}[!htb]
    \centering
    \includegraphics[width=0.90\textwidth]{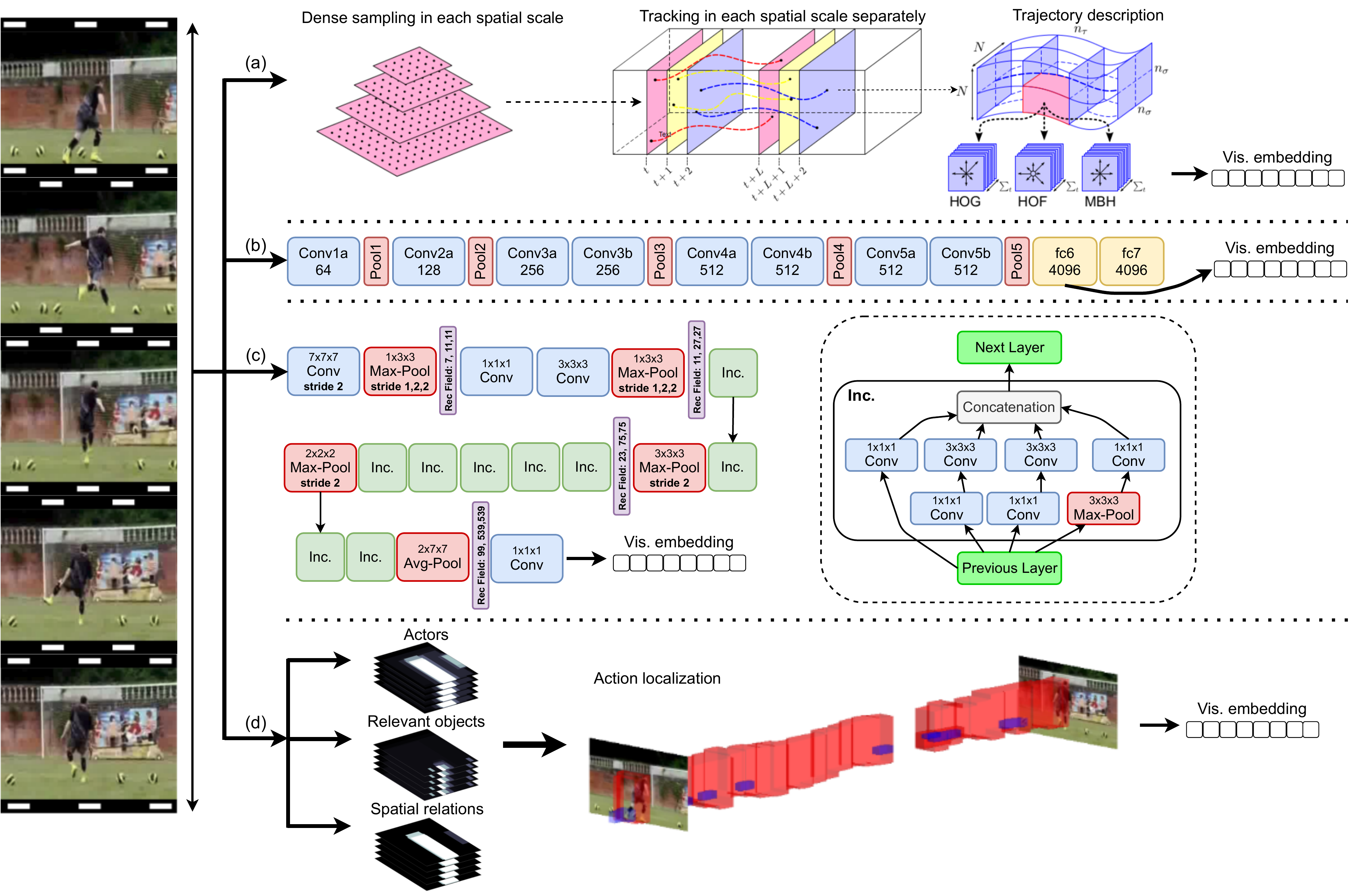}
    \caption{\textcolor{nblue}{Some visual embedding strategies that receive a common video clip and generate an array that represents global handcrafted features (a), deep features with temporal modeling ((b and c)), and actor-object relationships over the scene (d). The methods are (a) Dense trajectories~\citep{wang:2011}. (b) \gls{c3d}~\citep{tran:2015}. (c) \gls{i3d}~\citep{carreira:2017} and (d) Spatial-Aware Object Embeddings~\citep{mettes:2017-ref62}.}}
    \label{fig:visualembedding}
\end{figure*}

\textcolor{nblue}{In the visual embedding step, some methods, in most cases off-the-shelf, are used to process the visual information. These methods explore contiguous or sampled sequences of frames extracting global or local representations or identifying humans and objects and also how they interact with each other or evolve in the video clips. Figure~\ref{fig:visualembedding} illustrates a video segment processed with different methods.}

\textcolor{nblue}{We catalogue a set of methods used to perform visual embedding in the investigated literature, as shown in Table~\ref{tbl:visualembedding}. Next, we also provide a brief review of these methods.}

\gls{bof} methods are used in~\citep{rohrbach:2013-ref39,qiu:2011-ref11,liu:2011-ref58,rohrbach:2012-ref12,fu:2014-ref61}. 
In the first \gls{zsl} work in videos~\citep{liu:2011-ref58}, the visual words were obtained from a descriptor composed of spatio-temporal volumes and 1D Gabor detector. In later works, a well known combination of \gls*{hog}, \gls*{hof}, and \gls*{mbh} descriptors was used. 

However, more promising results were achieved with an improved \gls{bof} descriptor, \gls{dtf}, proposed in~\citep{wang:2011} and used in~\citep{xu:2015-ref19, guadarrama:2013-ref46}.
The \gls*{dtf} is able to characterize shape (point coordinates), appearance (\acrlong{hog}), motion (\acrlong{hof}) and variations on motion (\acrlong{mbh}). The \textit{dense} term refers to initial sampling in each frame with a grid of $W \times W$ points combined with spatio-temporal pyramid approach, as shown in Figure~\ref{fig:visualembedding}~(a) (on the left).

\begin{table}[!htb]
	\caption{Methods used to perform visual embedding in \gls{zsar} Handcrafted Features (HF) and Deep Features (DF).}
	\label{tbl:visualembedding}
	\centering
	\small
	\begin{tabular}{p{0.4cm}p{1.3cm}p{4.9cm}} \hline
		& Method                     &  Used in appoaches                                        \\ \hline
		HF    & \acrshort{bof}       & \citet{liu:2011-ref58}, \citet{qiu:2011-ref11},           \\
		&                            & \citet{fu:2014-ref61}, \citet{rohrbach:2013-ref39}, \\
		&                            &  \citet{rohrbach:2012-ref12}                                    \\
		\cline{2-3}               
		& \acrshort{dtf}             & \citet{xu:2015-ref19}, \citet{guadarrama:2013-ref46}      \\
		\cline{2-3}
		& \acrshort{idt}             & \citet{kodirov:2015-ref22}, \citet{gan2015-ref21},        \\
		&                            & \citet{xu:2017-ref20}, \citet{xu:2016-ref23},             \\
		&                            & \citet{xu:2017-ref24}, \citet{gan:2016-ref32},            \\
		&                            & \citet{alexiou:2016-ref54}, \citet{fu:2014-ref63} ,      \\ 
		&                            & \citet{zhang:2018-ref65}, \citet{liu:2018-ref26},         \\
		&                            & \citet{wang:2017-ref18}           \\
		\hline
		DF             & From~\citep{krizhevsky:2012} & \citet{jain:2015-ref31}                  \\    
		\cline{2-3}
		& VGG                        & \citet{gan:2016-ref32}, \citet{zhang:2018-ref65},         \\
		&                            & \citet{zuxuan:2016-ref71}                                 \\
		\cline{2-3}
		& ResNet-50                  & \citet{bishay:2019-ref78}                                 \\
		\cline{2-3}               
		& ResNet-200                 & \citet{zhu:2018-ref55}                                    \\
		\cline{2-3}               
		& \acrshort{3dcnn}           & \citet{mishra:2018-ref25}                                 \\
		\cline{2-3}               
		& \acrshort{c3d}             & \citet{wang:2020-ref15}, \citet{liu:2018-ref26},        \\
		&                            & \citet{zhang:2018-ref28}, \citet{hahn:2019-ref43},          \\
		&                            & \citet{wang:2017-ref14}, \citet{bishay:2019-ref78},       \\
		&                            & \citet{mandal:2019-ref72}, \citet{mishra:2020-ref73},     \\
		&                            & \citet{brattoli:2020-ref75}     \\
		\cline{2-3}               
		& \acrshort{i3d}             & \citet{roitberg:2018-ref56}, \citet{ghosh:2020-ref74}, \\
		&                            & \citet{mandal:2019-ref72}, \citet{piergiovanni:2018-ref57}        \\
		\cline{2-3}
		& R(2+1)D                    & \citet{brattoli:2020-ref75}     \\
		\cline{2-3}               
		& Other                      & \citet{li:2016-ref64}            \\
		\hline
	\end{tabular}
\end{table}

Since it is not possible to apply tracking in homogeneous regions of video frames, these points are removed from sampling. For each remaining point in each frame, the dense flow field is computed, and subsequent frames are concatenated to create a trajectory descriptor. Next, static trajectories of each sampled point are also removed (Figure~\ref{fig:visualembedding}~(a) (center)). Then, descriptors are computed from spatio-temporal volumes with $N \times N \times L$ dimension (e.g., 5 pixels $\times$ 5 pixels $\times$ 15 frames), subdivided in $n_\sigma \times n_\sigma \times n_\tau$ cells (e.g., 2 $\times$ 2 $\times$ 3), as shown in Figure~\ref{fig:visualembedding}~(a) (on the right). In the end, a codebook for each descriptor (trajectory, \gls{hog}, \gls{hof}, \gls{mbh}) is created by fixing the number of visual words per descriptor to 4000 and performing $k$-means algorithm eight times, while keeping the results with the lowest error. The resulting histograms of visual words are used as a global video representation.

As shown in~\citep{wang:2013b}, the performance of the \gls{hof} descriptor degrades significantly in the presence of camera motion (e.g., pan, tilt and zoom). Hence, the \gls{idt} method~\citep{wang:2013a} provides a mechanism to cancel out the camera motion from optical flow in the tracking phase, and a human detector~\citep{prest:2012} is used to remove trajectories in regions where humans are not found. \textcolor{nblue}{This method presents a promising performance and is used in many works (see Table~\ref{tbl:visualembedding})}. However, it is computationally intensive and becomes impracticable on large-scale datasets~\citep{tran:2015, liu:2018-ref26}.

\textcolor{nblue}{Deep learning has attracted much attention in recent years due to its advances in several problems such as: image classification~\citep{pouyanfar:2018}, hand gesture recognition~\citep{kopuklu:2019}, licence plate recognition~\citep{laroca:2018}, and spoofing detection~\citep{menotti:2015})}. In these applications, it is common to employ deep models pre-trained in large-scale datasets, and this ability is the major motivation for their use in \gls{zsar}. For example, a \gls{cnn} pre-trained on the ImageNet dataset~\citep{deng:2009}, called VGG 19~\citep{simonyan:2015}, is used in~\citep{gan:2016-ref32}, providing a detector for 1000 different concepts from individual frames. In their work, videos are represented in terms of detected visual concepts that are classified as relevant or irrelevant according to their similarity with a given textual query. \citet{jain:2015-ref31} also proposed an approach that relates objects and actions using the ImageNet dataset for training a \gls{cnn} model from~\citep{krizhevsky:2012}. In~\citep{zhu:2018-ref55}, a ResNet-200 model is initially trained on ImageNet and fine-tuned on ActivityNet dataset~\citep{heilbron:2015}. However, such image-based deep models are not suitable for direct video representation due to the lack of motion modeling, as demonstrated in~\citep{tran:2015}. This problem can be overcome with deep models that consider spatio-temporal relations, providing features from their fully connected layers (\texttt{fc}). This strategy is applied in~\citep{mishra:2018-ref25} using \gls{3dcnn}~\citep{ji:2013}, in~\citep{wang:2020-ref15, zhang:2018-ref28, liu:2018-ref26, hahn:2019-ref43, wang:2017-ref14} using \gls{c3d}~\citep{tran:2015}, and in~\citep{roitberg:2018-ref56, piergiovanni:2018-ref57} using \gls{i3d}~\citep{carreira:2017}.

In the \gls{c3d} network~\citep{tran:2015}, full video frames are taken as input and do not require any preprocessing except to resize frames to 128$\times$171 pixels. To propagate spatio-temporal information across all the layers, 3D convolutional filters (3$\times$3$\times$3 with stride 1$\times$1$\times$1) and 3D polling layers (2$\times$2$\times$2 with stride 2$\times$2$\times$2) are used. The architecture has two fully connected layers and a softmax output layer, which is removed to extract the visual embedding representation, as shown in Figure~\ref{fig:visualembedding}~(b). This model is trained on Sports-1M Dataset~\citep{karpathy:2014} and the visual representation is extracted from \texttt{fc6} layer resulting in a vector with 4096 dimensions which are usually used without modifications or fine-tuning. An exception occurs in~\citep{zhang:2018-ref28}, in which the dimensionality is reduced to 500 using \gls{pca}.

Training 3D \gls{cnn} consists of learning many more parameters than 2D \gls{cnn}.
Therefore, the \gls{i3d} architecture~\citep{carreira:2017} (Figure~\ref{fig:visualembedding}~(c)) uses a common pre-trained ImageNet Inception-V1 model~\citep{ioffe:2015} as base network, adding a batch normalization to each convolution layer. 
To properly explore spatio-temporal ordering and long-range dependencies, it uses a \gls{lstm} layer after the last average pooling layer of the Inception-V1. Additionally, its performance can be improved by including an optical-flow  stream~\citep{carreira:2017}. This model is shown in Figure~\ref{fig:visualembedding}~(c). The \gls{i3d} model is trained on Kinetics dataset~\citep{carreira:2017}, and the visual representation is extracted from the last fully connected layer resulting in a representation of 256 dimensions in~\citep{roitberg:2018-ref56} and 1024 in~\citep{piergiovanni:2018-ref57}. It is likely that \gls{zsl} assumption (classes disjunction between the training and testing sets) has been violated since both \gls{c3d} and \gls{i3d} models are pre-trained on large-scale datasets~\citep{liu:2018-ref26}. 
Thus, a new problem emerges through the use of deep learning techniques. \textcolor{nblue}{We present a detailed discussion on this topic in Section~\ref{sec:performance}}. Although simple and relatively effective, recent works have shown significantly gain in performance when these off-the-shelf global descriptors are fine-tuned, used to model temporal or spatial relationships, or conditioned by semantic information to produce new representations.

\subsection{Semantic Label Embedding Step}
\label{sec:labelembedding}

\begin{figure*}[!htb]
     \centering
     \begin{subfigure}[b]{0.58\textwidth}
         \centering
         \includegraphics[width=\textwidth]{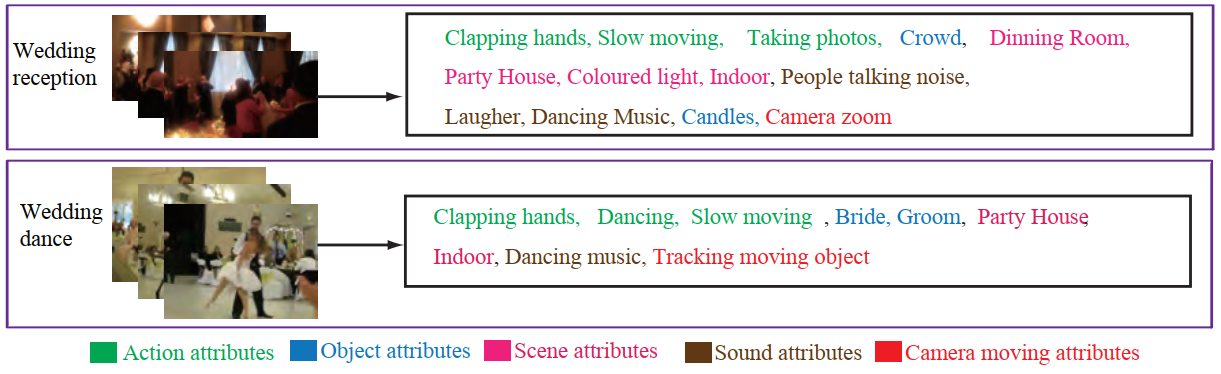}
         \caption{}
         \label{fig:semembedd_a}
     \end{subfigure}
     \hfill
     \begin{subfigure}[b]{0.31\textwidth}
         \centering
         \includegraphics[width=\textwidth]{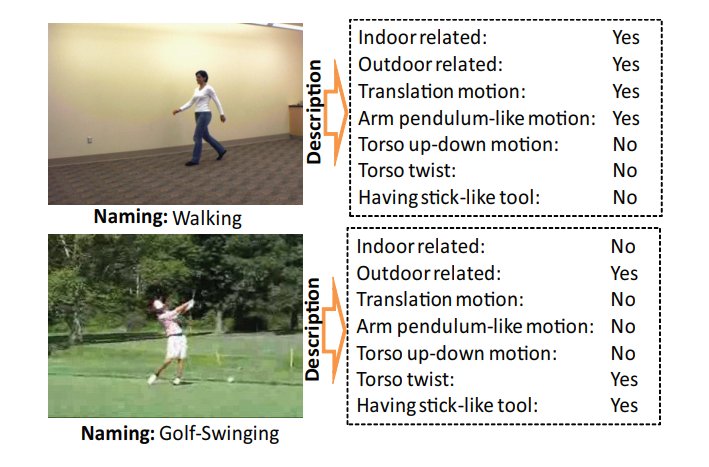}
         \caption{}
         \label{fig:semembedd_b}
     \end{subfigure}
     \hfill
     \linebreak
     \begin{subfigure}[b]{0.25\textwidth}
         \centering
         \includegraphics[width=\textwidth]{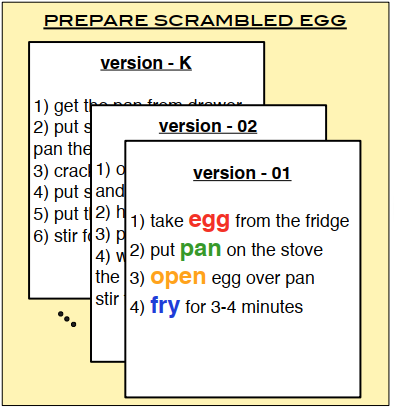}
         \caption{}
         \label{fig:semembedd_c}
     \end{subfigure}
     \hfill
     \begin{subfigure}[b]{0.38\textwidth}
         \centering
         \includegraphics[width=\textwidth]{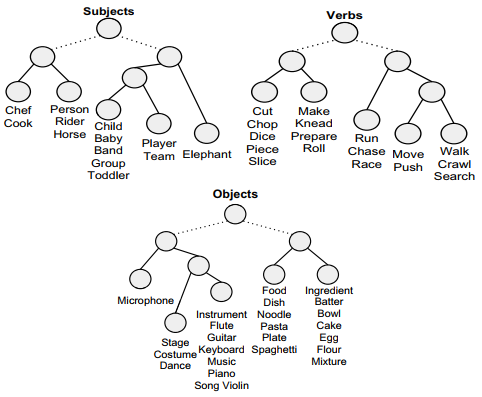}
         \caption{}
         \label{fig:semembedd_d}
     \end{subfigure}
     \hfill
     \begin{subfigure}[b]{0.22\textwidth}
         \centering
         \includegraphics[width=\textwidth]{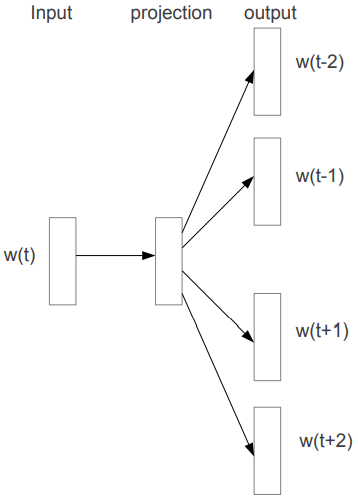}
         \caption{}
         \label{fig:semembedd_e}
     \end{subfigure}
     \hfill
        \caption{\textcolor{nblue}{Main strategies for performing semantic label embedding in \gls{zsar}.
        (a) The methods proposed by~\citet{fu:2012-ref48} and (b) \citet{liu:2011-ref58} are attribute-based. (c) The approach developed by~\citet{rohrbach:2012-ref12} is a script-data representation. (d) The scheme proposed by~\citet{guadarrama:2013-ref46} is a semantic hierarchy; (e) The approach developed by~\citet{mikolov:2013-b} is an unsupervised word embedding method.}}
        \label{fig:semanticembedding}
\end{figure*}

Providing meaningful semantic information in \gls{zsar} is a challenging task. On one hand, we can utilize attribute-based approaches, that have several drawbacks, such as: (i) annotating videos is more difficult than annotating images; (ii) in a more complex or complete dataset, several attributes are necessary to alleviate the semantic intraclass variability; (iii) it is difficult to define what attributes are relevant, and (iv) this approach is not scalable. \textcolor{nblue}{On the other hand, we can utilize textual corpus information, which typically relies on exploring unsupervised word embedding methods, gaining with a scalable process but losing performance. Figure~\ref{fig:semanticembedding} illustrates some strategies to perform semantic embedding. Next, we detail the main approaches.}

\begin{table}[!htb]
	\caption{\textcolor{nblue}{Methods used to perform semantic embedding in \gls{zsar}. Attribute (A) and Word Embedding (WE).}} 
	\label{tbl:semanticembedding}
	\centering
	\small	 
	\color{nblue} 
	\begin{tabular}{p{0.4cm}p{1.4cm}p{4.8cm}} \hline
		& Method                         &  Used in appoaches                                     \\ \hline
		A     & Annotated                & \citet{wang:2017-ref18}, \citet{fu:2014-ref63},    \\
		&                                & \citet{mishra:2018-ref25}, \citet{liu:2011-ref58},         \\
		&                                & \citet{rohrbach:2013-ref39}, \citet{gan2015-ref21},    \\
		&                                & \citet{fu:2014-ref61}, \citet{bishay:2019-ref78},      \\
		&                                & \citet{mandal:2019-ref72}, \citet{mishra:2020-ref73}   \\
		\cline{2-3}               
		& Dictionary learning            & \citet{qiu:2011-ref11}, \citet{kodirov:2015-ref22}     \\
		\cline{2-3}
		& Dynamic                        & \citet{jones:2019-ref76}      \\
		\hline
		WE       & Semantic hierarchies  & \citet{rohrbach:2012-ref12}, \citet{gan2015-ref21}     \\ 
		\cline{2-3}
	    & Knowledge graphs               & \citet{ghosh:2020-ref74}, \citet{gao:2019-ref70}       \\
		
		\cline{2-3}               
		& Word2Vec                       & \citet{xu:2015-ref19}, \citet{xu:2017-ref20},          \\
		&                                & \citet{xu:2016-ref23}, \citet{jain:2015-ref31},        \\
		&                                & \citet{gan:2016-ref32}, \citet{alexiou:2016-ref54},    \\
		&                                & \citet{li:2016-ref64}, \citet{zuxuan:2016-ref71},      \\
		&                                & \citet{xu:2017-ref24}, \citet{wang:2020-ref15},        \\
		&                                & \citet{qin:2017-ref35}, \citet{mishra:2018-ref25},        \\
		&                                & \citet{wang:2017-ref18}, \citet{liu:2018-ref26},    \\
		&                                & \citet{brattoli:2020-ref75}, \citet{zhu:2018-ref55},     \\
		&                                & \citet{roitberg:2018-ref56}, \citet{mandal:2019-ref72}, \\
		&                                & \citet{bishay:2019-ref78}, \citet{roitberg:2018-ref65},  \\
		&                                & \citet{hahn:2019-ref43}, \citet{mishra:2020-ref73},    \\
		&                                & \citet{mettes:2017-ref62}   \\
		\cline{2-3}                         
		& GloVe                          & \citet{zhang:2018-ref28}, \citet{piergiovanni:2018-ref57},                                                      \citet{zhang:2018-ref65}, \citet{guadarrama:2013-ref46} \\  
		\hline               
	\end{tabular}
\end{table}

As shown in Table~\ref{tbl:semanticembedding}, using manually defined and annotated attributes is a common strategy~\citep{wang:2017-ref18, mishra:2018-ref25, fu:2014-ref63, liu:2011-ref58, rohrbach:2013-ref39, gan2015-ref21, fu:2014-ref61}. An expert needs to define all attributes and also their values. 
\textcolor{nblue}{These annotations can be made directly (e.g., annotations from UCF101, or USAA~(Figure~\ref{fig:semembedd_a})); or acquired by processing textual descriptions in the form of script-data (Figure~\ref{fig:semembedd_c}, collected with \gls{amt}, as described by~\citet{rohrbach:2012-ref12}. Alternatively, dictionary learning techniques are proposed in~\citep{qiu:2011-ref11, kodirov:2015-ref22}. In these works, visual features are related to atoms in the automatically learned dictionary, alleviating the problem of manual definition of attributes.}

Recently, methods based on word embedding have become popular (Table~\ref{tbl:semanticembedding}). For example, semantic hierarchies are mined for subjects, verbs, and objects using the descriptions of videos from YouTube~\citep{guadarrama:2013-ref46} (Figure~\ref{fig:semembedd_d}) and WordNet~\citep{fellbaum:1998} hierarchy was used by~\citet{rohrbach:2012-ref12} to represent the action labels. 
\textcolor{nblue}{However, the most popular strategy for semantic label embedding is the skip-gram model~\citep{mikolov:2013-a} (Figure~\ref{fig:semembedd_e}), more specifically the Word2Vec implementation~\citep{mikolov:2013-b} used in a wide variety of works (see Table~\ref{tbl:semanticembedding}). This model is an efficient method for learning vector representations that captures a large number of syntactic and semantic word relationships~\citep{mikolov:2013-b}.} The method consists of learning a neural network that calculates a similarity measure between words based on a softmax output. In \gls{zsl}, the semantic vector representation for the interest word (action label) is based on the activation of 300 neurons in a hidden layer of the skip-gram network when this word is provided as input. 

\begin{figure*}[!htb]
     \centering
     \begin{subfigure}[b]{0.45\textwidth}
         \centering
         \includegraphics[width=\textwidth]{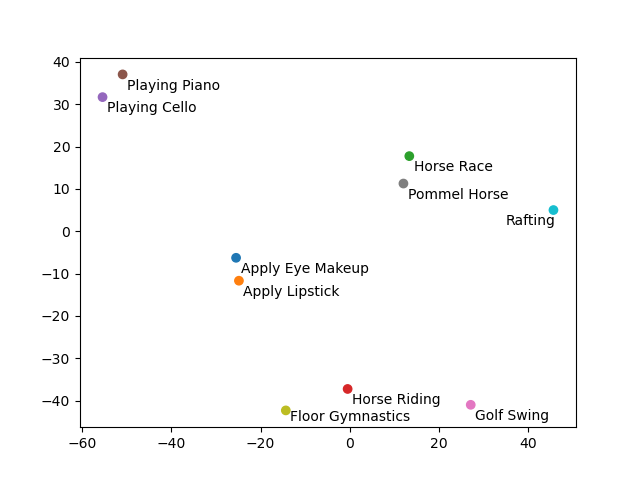}
         \caption{}
         \label{fig:word2vecembeddings}
     \end{subfigure}
     \hfill
     \begin{subfigure}[b]{0.45\textwidth}
         \centering
         \includegraphics[width=\textwidth]{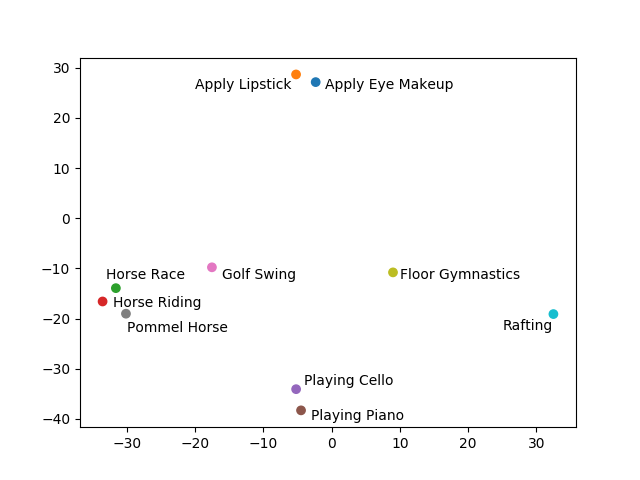}
         \caption{}
         \label{fig:gloveembeddings}
     \end{subfigure}
     \hfill
        \caption{\textcolor{nblue}{Word embeddings of 10 classes from UCF101 using in (a) Word2Vec~\citep{mikolov:2013-b} and (b) GloVe~\citep{pennington:2014} methods. In both cases, the original representations have their dimensionality reduced using t-sne~\citep{maaten:2008}.}}
        \label{fig:semanticembeddingspace}
\end{figure*}

Another approach to performing semantic label embedding is a count-based model called Global Vectors (GloVe)~\citep{pennington:2014}. In that model, a large matrix of co-occurrence statistics is constructed by storing words in rows and contexts in columns. Semantic vectors are learned such that their dot product equals the co-occurrence probability~\citep{akata:2014}. Intuitively, these statistics encode the meaning of words since the frequency of semantically similar words is higher than semantically dissimilar words. 
\textcolor{nblue}{This word embedding property can be observed in Figure~\ref{fig:semanticembeddingspace} with the class pairs \texttt{Playing Cello-Playing Piano}, \texttt{Apply Eye Makeup-Apply Lipstick} in both Figures~\ref{fig:word2vecembeddings} and~\ref{fig:gloveembeddings}. In these figures, 10 class representations from UCF101, acquired with Word2Vec and GloVE, were projected onto 2-dimensional spaces using the t-sne method~\citep{maaten:2008}. We adopt a general approach to combine two or more word embeddings with a simple average of them. This approach is efficient but, in some cases, produces semantic imprecision as in the cases of \texttt{Horse Race-Horse Riding}, distant from each other in~\ref{fig:word2vecembeddings}, or \texttt{Pommel Horse-Horse Race} close to each other. Therefore, strategies based on textual descriptions or action-object relationships are successful and expected in future works.}

\section{Zero-Shot Action Recognition Approaches}
\label{sec:trainingstep}

The central problem in \gls{zsar} is how to use visual and semantic information to classify new instances from unseen classes (i.e., to perform transfer knowledge). We identify three main approaches: (i) to classify directly into the semantic embedding space, usually projecting the visual features on it; 
(ii) to classify into an intermediate space generated with some combination technique for both visual and semantic representation (e.g., latent attributes or co-occurrence of actions and objects) and; (iii) to classify into the visual embedding space by synthesizing visual features conditioned by semantic side information in order to produce visual prototypes for unseen classes. Many other taxonomies could be proposed. \textcolor{nblue}{However, \gls{zsar} methods combine multiple strategies, so that provide unambiguous classifications is very difficult. Table~\ref{tbl:zsaroverview} presents the methods and their classification according to our general criteria, and we also provide some observations}.

In the next subsections, we explain the general ideas of these methods with a common notation and avoiding math complications whenever possible. In \gls{zsar}, there are two datasets: the first is the training dataset $D_{tr} = \{(x_{n}, y_{n})\}_{n = 1}^{N_{s}}$, and the second is the testing dataset $D_{te} = \{(x_{n}, y_{n})\}_{n=1}^{N_{u}}$, where $x_{n}$ and $y_{n}$ are, respectively, the visual representation and the class label for the n\textit{-th} video sample $v_{n}$, $N_{s}$ is the number of seen examples, and $N_{u}$ is the number of unseen examples. The label spaces are $\mathcal{S}=\{1,2,...,S\}$ and $ \mathcal{U}=\{S+1,S+2,...,U\}$ with $\mathcal{S} \cap \mathcal{U} = \emptyset$. The visual feature is embedded with a function $E_{v}(v_{n})= x_{n}$ so that $x_n \in R^{d}$. As discussed earlier, the function $E_{v}$ may be represented by the methods \gls{dtf}, \gls{idt} or \gls{c3d}, for example, as illustrated in Figure~\ref{fig:visualembedding}. Similarly, the semantic embedding function for each class label is $E_{l}(y_{n})= z_{n}$ so that $z_{n} \in R^{m}$. \textcolor{nred}{The $E_{l}$ function usually corresponds to manual attribute annotation, data-driven attributes, learned hierarchies or word vectors, as shown in Figure~\ref{fig:semanticembedding}}.

\begin{table*}
	\caption{\textcolor{nblue}{Overview of zero-shot human action recognition methods in videos. We organize the methods into three categories: classification into the semantic space, classification into an intermediate space, and classification into the visual space. For each approach, we point out the main strategies adopted.}}
	\label{tbl:zsaroverview}
	\centering
	\scriptsize
	\color{nblue}
	\resizebox{0.95\textwidth}{!}{
	\begin{tabular}{p{3.5cm}m{7.2cm}} \hline
		
		\multicolumn{2}{c}{Classification into the semantic space} \\ \hline
		Reference &   Main strategies             \\\hline
		\citet{liu:2011-ref58}      & Single task learning with support vector machine   \\
		\citet{fu:2012-ref48}       & PTM + LDA + NN                                     \\
        \citet{fu:2014-ref61}       & PTM + LDA + NN + unconstrained attribute learning  \\
		\citet{qiu:2011-ref11}      & Sparce dictionary learning                         \\
		\citet{rohrbach:2012-ref12} & Text score + NN or SVM                             \\
		\citet{rohrbach:2013-ref39} & Text score + NN or SVM + transductive setting      \\
		\citet{xu:2015-ref19}       & Non-linear SVR with kernel RBF-$_{\chi^{2}}$       \\
		\citet{xu:2017-ref24}       & Manifold RR + transd. setting + data augmentation  \\
		\citet{kodirov:2015-ref22}  & Dictionary learning + regularised sparse coding    \\
		\citet{li:2016-ref64}       & MLP + convex combination of similar embedding      \\
		\citet{hahn:2019-ref43}     & Temporal modeling with LSTM + verb relationships   \\
		\citet{alexiou:2016-ref54}  & Semantic improvements with synonyms + self training \\ 
		\citet{bishay:2019-ref78}   & Relation networks with segment-by-segment attention \\
		\citet{brattoli:2020-ref75} & End-to-end model with linear classifier + cross-dataset \\\hline
		
		\multicolumn{2}{c}{Classification into an intermediate space} \\ \hline
		Reference &   Main strategies             \\\hline
		
		\citet{xu:2016-ref23}           & Multi-task learning + prioritized data augmentation \\
		\citet{fu:2014-ref63}           & Multi-view embedding space + CCA \\
		\citet{wang:2017-ref18}         & Landmark-based learning + sammon mapping + ST + SP \\
		\citet{wang:2017-ref14}         & Exploring texts and images for semantic embedding \\
		\citet{wang:2020-ref15}         & Multi-label ZSL + new split scheme \\
		\citet{gan2015-ref21}           & Concept detectors using least square regression (LR) \\
		\citet{zhu:2018-ref55}          & Universal representation with GMIL + NMF \\
		\citet{mishra:2018-ref25}       & Linear combinations of basis vectors (Gaussian params) \\
		\citet{mishra:2020-ref73}       & Synthesized features with IAF and bi-di GAN \\
		\citet{qin:2017-ref35}          & Visual embedding with error correcting output codes \\
		\citet{guadarrama:2013-ref46}   & Semantic hierarchies for subjects, objects and verbs \\
		\citet{jain:2015-ref31}         & Affinity between objects and classes \\
		\citet{zuxuan:2016-ref71}       & Semantic fusion network for objects, scenes and actions \\
		\citet{poppe:2010-ref62}        & Spatial-aware object embedding in action tubes \\
		\citet{gao:2019-ref70}          & Action-object relationship modeled with GCN \\
		\citet{zhang:2018-ref28}        & Multi-modal learning using video and text pairing \\
		\citet{piergiovanni:2018-ref57} & Video and text encoding with unpaired data \\ 
		\citet{jones:2019-ref76}        & Dynamic attributes signatures + finite state machines \\
		\citet{ghosh:2020-ref74}        & Knowledge graphs learning + GCN \\\hline
		
		\multicolumn{2}{c}{Classification into the visual space} \\ \hline
		Reference &   Main strategies             \\\hline
		
		\citet{zhang:2018-ref65}  & Joint distribution of visual and semantic knowledge \\
		\citet{mandal:2019-ref72} & Synthesized features + out-of-distribution classifier \\
		                  
		\hline
	\end{tabular}}
	\end{table*}

\subsection{Classification into the semantic embedding space}

Many works try to learn a function $p: \mathbf{x} \xrightarrow{} \mathbf{z}$ to project a visual representation $x_{n}$ onto the semantic space obtaining a $z'_{n}$ representation. Then, a function $q: \mathbf{z'} \xrightarrow{} \mathbf{y}$ is learned. In most cases this $q$ function is a simple nearest neighbor classifier. 
However, \citet{liu:2011-ref58} proposed a latent \gls{svm} formulation for the $p$ function. In this case, $m$ individual attribute classifiers $(p_{m})$ maps each representation $x$ to the \textit{i}-th correspondent attribute of $z$ (i.e., each dimension). They do not use only annotated attributes, but also learn data-driven attributes by clustering low-level features maximizing the system information gain and using these features as latent variables. An unseen instance is classified using the $p_{m}$ functions to project the raw visual features onto the semantic space and performing a nearest neighbour classification with Euclidean distance.

\citet{fu:2012-ref48} introduced an attribute learning method based on the \gls{ptm} with \gls{lda}~\citep{blei:2003}. In~\citep{fu:2014-ref61}, they present the \gls{m2latm} that extends that prior method. The new formulation considers three types of attributes: user-defined (UD), from any prior ontology (e.g., USAA dataset), latent class-conditional (CC), discriminative for known classes, and generalized free (GF), which represents shared aspects not presented in the attribute ontology. The major difference to~\citep{fu:2012-ref48} is the adaptation of \gls{ptm} to work unconstrained. For example, UD topics are constrained in 1 to 1 correspondence with attributes from the ontology, and latent CC topics are constrained to match the class label. On the other hand, GF attributes are unconstrained. In both works, the classification occurs using a nearest neighbor rule using cosine distance. Another similar work appears in~\citep{qiu:2011-ref11}. In this case, the $p$ function corresponds to a dictionary learning method via information maximization. Both appearance information between dictionary atoms and class label information are combined in order to learn a compact and discriminative dictionary for human action attributes.

Some approaches use only predefined semantic attributes. For example, \citet{rohrbach:2013-ref39} proposed to associate scores of visual features with semantic attributes based on visual-annotation alignment, contextual, and co-occurrence information of the attributes. In their work, basic-level cooking actions such as fry or open and their related objects, egg or pan, are taken as attributes. Side information from cooking scripts (i.e., script-data shown in Figure~\ref{fig:semembedd_c}) is used to select the most relevant attributes based on statistical scores (frequency and term frequency $\times$ inverse document frequency) as in~\citep{rohrbach:2012-ref12}. The association between weighted attributes with the action label is learned using a nearest neighbor or a \gls{svm} classifier. They investigated a transductive setting by constructing a k-nearest neighbor ($k$-NN) graph calculating weights for instances in the semantic attribute space instead of the visual space. This approach exploits the manifold structure by utilizing the attributes from unknown classes without their class labels. \textcolor{nred}{The raw visual features (e.g., \gls{dtf}) are used to learn $m$ classifiers for each attribute, in a similar manner to~\citep{lampert:2009}. Then, the probability of new classes is estimated with script-data}.

Manual attributes have inherent limitations early discussed. \textcolor{nred}{Therefore, motivated by the success of the word vectors in language processing, many works try to extend their approaches to use this type of representation.} \citet{xu:2015-ref19} proposed an approach to project the low-level visual features obtained with \gls{mbh} and \gls{hog} onto a semantic space of 300-dimensions composed by word vectors from Word2Vec~\citep{mikolov:2013-a}. They trained a non-linear \gls{svr} with RBF-$_{\chi^{2}}$ kernel defined as
\begin{equation}
    K(x_{i}, x_{j}) = exp(-\gamma \cdot D(x_{i}, x_{j})),
    \label{eq:svr_ref19}
\end{equation}
\noindent where $D(x_{i}, x_{j})$ is the $\chi^{2}$ distance between histogram-based representations $x_{i}$ and $x_{j}$ to project new instances and classify with the nearest neighbour rule. \textcolor{nred}{Approaches similar to this formulation suffer severely with the domain shift problem because the probability distribution of seen classes is different from unseen ones. In their work, this problem is tackled with a transductive self-training procedure, and a data augmentation is conducted}.

Subsequently, in~\citep{xu:2017-ref20}, the authors proposed improvements in that approach using a manifold-regularized regression (semi-supervised learning). \textcolor{nred}{As observed by~\citet{dinu:2014}, in higher dimensional spaces, some instances from different classes may appear closest to each other, which is called the hubness problem}. To tackle this problem and leverage the accuracy, they adapted the manifold-regularized regression to explore the manifold structure of unseen classes in a transductive setting.

\textcolor{nred}{Simple projection methods do not treat suitably the differences between class distributions of seen and unseen datasets. Hence, they are prone to suffer from the domain shift problem.} To alleviate this problem, \citet{kodirov:2015-ref22} proposed an unsupervised domain adaptation model by regularized sparse coding. In their work, each dimension of the semantic embedding space corresponds to a dictionary basis vector $z_{i}$ with $m$ dimensions. If the visual features are represented by a vector $\mathbf{x_{i}}$ with $d$ dimensions, a dictionary $D^{d \times m}$ can be learned using quadratic optimization so that the reconstruction error of $\mathbf{x_{i}} = D\mathbf{z_{i}}$ is minimized. Two dictionaries are learned, one to the source dataset $D_{s}$ and another to the target dataset $D_{t}$. The domain shift is tackled by adding two constraints: $D_{t}$ should be similar to $D_{s}$ and, a visual-semantic similarity constraint given by the closeness of the interpretations of target data $z'_{i}$ to their true class prototype $z_{i}$. Once trained, $D_{t}$ is used to project the raw example onto the semantic space, and the nearest neighbor classifier assigns a label. Another strategy is to apply a label propagation across multiple semantic spaces, which are combined with a graph similarity matrix.

\citet{li:2016-ref64} proposed to learn a common embedding space using a \gls{mlp} to project visual features onto a 300-dimensional space where the class prototypes from Word2Vec are. The visual feature comes from a composition of two \gls{cnn} outputs. The first for appearance patterns (i.e., RGB flow) and the second for motion patterns (i.e., Optical Flow). The last fully connected layers of these models are combined and used as input to the \gls{mlp}. \textcolor{nred}{A new strategy to domain adaptation called \gls{conssev} is proposed. The main idea is to adjust the semantic output from \gls{mlp} by creating a new vector weighted by the sum of all $k$ highest similar vectors}. This similarity is given by the \gls{mlp} softmax outputs.

\textcolor{nred}{A similar strategy is presented in~\citep{hahn:2019-ref43}, but including temporal modeling.} In their work, the videos are represented in a scheme in which the visual features are slightly related to corresponding verbs represented as word embeddings from Word2Vec method. In this method, until 21 short clips per video are fed to a \gls{c3d} model obtaining 21 vectors with 4096 dimensions. After, these vectors are grouped into 7 groups of 3 vectors each and used to train a network composed by 2-layer \gls{lstm} units and a fully-connected layer with 300 dimensions. The network is trained with a loss function defined as a sum of a cross-entropy loss ($L_{CE}$) and a pairwise-ranking loss ($L_{PR}$) defined as
\begin{equation}
\begin{split}
L_{PR} =  \min_{\theta}{\sum_{i}\sum_{x}(1 - s(a_{i}, v_{i}))} +\\\max\{0, s(a_{x}, v_{i})\} + \max\{0, s(a_{i}, v_{x})\},
\end{split}
\label{eq:paired_loss_function_ref43}
\end{equation}
\noindent where $s$ is a similarity function (e.g., cosine similarity), $v_{i}$ is a verb embedding of a class $i$ (i.e., their word vector), $a_{i}$ is the embedding of action-video, $a_{x}$ is an action-video embedding of contrastive class $k$, and $v_{x}$ is a contrastive verb embedding of class $k$.
Zero-shot classification is performed by inputting a new example into the neural network and looking by the nearest neighbor of their 300-dimensional representation of the verbs into the semantic space. 

\citet{alexiou:2016-ref54} explored the impact of using class label synonyms on enhancing word vector representations. They mine a list of synonyms from multiple dictionaries for each class word vectors. 
These synonyms update the class word vector representations by weighting them based on the distances of actions and their synonyms. \textcolor{nred}{They also explored a self-training strategy, and the results using \gls{zsar} methods based on direct projection point out to accuracy improvements. However, the comparisons with other methods are difficult due to experimental protocol differences}.

\textcolor{nblue}{\citet{bishay:2019-ref78} proposed a method for \gls{fsl} that can be adapted to \gls{zsl}. Their main idea for \gls{fsl} is to estimate a deep similarity score among a query video and representative videos from each class assigning the label correspondent to the maximum score. In \gls{zsl}, this similarity is estimated among a query video and semantic embedding vectors representing the class labels. The model architecture in \gls{zsl} configuration has two modules: embedding and relation. The embedding module has two elements, the first one for visual embedding compound by a C3D network pre-trained on Sports-1M and the second one to semantic embedding compound by a skip-gram model. The relation module implements a segment-by-segment-attention mechanism that estimates the similarity between the semantic vector and each query video segment. The comparison outputs are aggregated over all segments using fully connected layers and an average pooling layer producing a final relation score. In the experiments, they adopt 50\%/50\% and 80\%/20\% random splits with 30 trials. They did not inform if overlapping classes between their test set and the training set used in the pre-training deep model (C3D) were removed. Therefore, the results can violate the \gls{zsl} restriction.}

\textcolor{nblue}{More recently, \citet{brattoli:2020-ref75} proposed the first end-to-end approach in \gls{zsar}. In their method, both visual embedding and semantic embedding are learned and optimized at a same time. The visual embedding is acquired using $R(2+1)D$~\citep{tran:2018} or C3D~\citep{tran:2015} architectures. The output of these models is $B\times T \times$512, where $B$ is the batch size and $T$ the number of clips (1 in training and 1 or more in testing). $E_{s}$ is a linear classifier with 512$\times$300 weights, and, therefore, the output of $E_{s} \circ E_{v}$ is of shape $B\times300$. A Word2Vec is incorporated given representations for all labels with 300-d. The loss function adopted consists of minimizing Equation~\ref{eq:loss-ref75}.}
\begin{equation}
    \color{nblue}
    L = \sum{\parallel W2V(c) - (E_{s} \circ E_{v})(x^{t}) \parallel^{2}}
    \label{eq:loss-ref75}
\end{equation}

\textcolor{nblue}{Their work adopts a more realistic scenario cross dataset, where no overlapping between seen and pre-training classes are required to preserve the ZSL restriction. The similarity evaluation between class labels follows the protocol proposed by~\citet{roitberg:2018-ref65}, where a label must be conveniently distant from any class label used to train the model.}

\subsection{Classification into an intermediate space}

The techniques surveyed in this subsection create an intermediate space by projecting both visual embedding $E_{v}$ and the semantic embedding $E_{s}$ onto a new common $t$-dimensional space $q$, where $q \in R^{t}$. Typically, the visual projection occurs such as in the direct projection approaches, such that is necessary to develop methods suitable to project the semantic features onto this new subspace, which is the main focus of these approaches.

\citet{xu:2016-ref23} proposed a \gls{mtl} approach instead of learning $m$ classifiers (e.g., single task ridge regression \citep{xu:2017-ref20}). \textcolor{nred}{They argue that single-task leads to overfitting because they assume each dimension independently, disregarding their relationships.} With \gls{mtl}, the parameters of all tasks lie on a low dimensional manifold. They also proposed a prioritized auxiliary data augmentation\footnote{From multiple domains.} for domain adaptation by selecting the most relevant instances for each class by minimizing the discrepancy between the marginal distributions of the auxiliary and target domains. \textcolor{nred}{This procedure is important because it may occur negative transfer learning due to the dissimilarity between the extra incorporated data and the target classes for recognition.} More specifically, they generalize the \gls{kliep} for zero-shot learning problem providing a vector with weights $\mathbf{w}$ that are applied to $\mathbf{x}$ jointly with \gls{mtl} to create an intermediate space.

\citet{fu:2014-ref63} proposed a transductive multi-view embedding space to alleviate the domain shift problem. To build this latent joint space, they extracted low-level features and projected this representation onto the semantic spaces of multi-view sources (i.e., attributes and word vectors in this case) using single task classifiers (e.g., the same used in~\citep{liu:2011-ref58} or~\citep{xu:2017-ref20}). The vector spaces are combined with \gls{cca} in order to find linear combinations between the semantic vectors by maximizing the correlation among the attributes. 
They utilized the eigenvalues of each dimension as a weight estimator that highlights some characteristics for each class. The zero-shot classification is leveraged with a heterogeneous hypergraph-based semi-supervised learning used to explore the manifold structure of the unlabelled data transductively.

\citet{wang:2017-ref18} proposed a method based on two stages, i.e., BiDiLEL. In the first stage, a latent embedding space is first created, learning a projection function that maps the visual features onto this low-dimensional subspace. A class landmark is calculated as a mean representation of all instances of that class. In the second stage, an adaptation of Sammon mapping~\citep{sammon:1969} is proposed, called landmark-based Sammon mapping~(LSM), responsible for projecting the semantic representation onto the latent space preserving the semantic relatedness between all different classes using the landmarks as guides. The \gls{zsl} classification consists of extracting visual features, projecting the representation onto latent space and, searching for the nearest landmark neighbor. Additionally, techniques for post-processing such as self-training and structured prediction were used. Posteriorly, using the BiDiLEL method, \citet{wang:2017-ref14} studied different semantic representations for bridging the semantic gap. Their alternative representations are based on textual descriptions of human actions and deep features extracted from still images relevant to human actions. For textual-based descriptions, a corpus obtained from the Wikihow, Wikipedia and Online dictionary is preprocessed with natural language techniques (e.g., obtaining all words in documents and removing stopping words such as ``is'', ``you'', ``of''). The word vectors are represented as average word vectors or Fisher word vectors. On the other hand, for image-based description, a dataset is created using action labels as keywords and relevant images collected with search engines. \textcolor{nred}{These images are inputted into a pre-trained \gls{cnn} model, where the resultant deep image features are coded as average feature vectors or Fisher feature vectors, resulting in higher performance.}

They also investigated the multi-label \gls{zsl} problem in~\citep{wang:2020-ref15} based on the observation that, in real scenarios, a video clip conveys multiple human actions corresponding to different concepts and then proposed a multi-label classification method based on a joint ranking embedding learning. \textcolor{nred}{However, their main contribution is a novel data split designed specially to this problem.} Instead of using a usual instance-first split, they proposed a label-first split in which all the labels are first divided into two mutually exclusive subsets (i.e., seen and unseen). Next, instances that have at least one unseen label are kept for testing, and the rest is taken as seen labels.
\textcolor{nred}{Hence, the seen subset may be divided into training and validation splits, suitably simulating the real world \gls{zsl} scenario.}

The method proposed by~\citet{gan2015-ref21} considers that action classes may share some elements if they are semantically similar to each other. The visual representation is used to learn concept detectors for each class by applying \gls{lr} as
\begin{equation}
    arg_{w_{k}} \sum_{n}(w_{k}^{T}x_{i}-y_{i})^{2}+\lambda\parallel w_{k} \parallel ^{2}
    \label{eq:gan1}
\end{equation}
\noindent where $x_{i} \in R^{d}$ is the low-level feature for a video $i$ and $y_{i} \in \{0, 1\}$ is the associated binary label to the class $k$. In practice, $w$ is interpreted as the concept detector and $\lambda$ is a regularization term. The authors explored different values for it (e.g., 0.01, 0.1, 1, 10, 100). They utilized the WordNet hierarchy and the Word2Vec model to infer the semantic similarity between the class labels with a function from \citet{lin:1998} and the cosine distance, respectively. Thus, the classification problem can be expressed as
\begin{equation}
    p(y_{u}|x) = \sum_{k = 1}^{K}p(y_{u}|y_{k})p(y_{k}|x),
    \label{eq:gan2}
\end{equation}
\noindent where $p(y_{u}|x)$ is the \textit{a posteriori} zero-shot classification probability (for the class $y_u$ given $x$), $p(y_{u}|y_{k})$ is given by semantic similarity from side information and the concept detectors. The $p(y_{k}|x)$ is calculated with the concept detectors. \textcolor{nred}{Although their promising reported performance, their work only splits the dataset into 90\% for seen and 10\% for unseen classes, and an evaluation of how the reduced number of seen classes affects the accuracy of the concept detectors is not provided. We believe that performance will be strongly degraded if only 50\% of classes were taken as seen.}

The main idea of \citet{zhu:2018-ref55} was to find the most relevant basis to discriminate an action. Then, they combined this information with semantic word embedding to create a generic representation for actions called \gls{ur}. UR is computed with \gls{gmil} by evaluating if one instance is more attractive or repulsive to the action class patterns and joining the first ones in bags with pooled Naive Bayes Nearest Neighbor. The \gls{ur} consists in correlating visual features with semantic information (e.g., word vectors) in a common space $D_{s}: A \times B$, where $A = E_{v}(x_{s})$ is the visual embedding and $B = E_{s}$ is the semantic embedding. \gls{nmf} is employed to find two non-negative matrices from $A$ and another two to $B$ so that \gls{jsd} can be applied to preserve the generative components from \gls{gmil}, producing the \gls{ur}.
As this approach is focused on cross dataset problem, the domain shift is unknown. The strategy adopted is to use \gls{ur} to estimate the differences between the classes in the semantic modality. \textcolor{nred}{Therefore, using \gls{ur}, the misalignment observed with semantics are assumed to be reproduced in visual patterns, which is not always true.}

\citet{mishra:2018-ref25} proposed to represent the visual pattern of actions as a Gaussian distribution probability parameters $\mathbf{\mu_{c}}$ (mean vector) and $\mathbf{\sigma_{c}^{2}}$ (vector of diagonal covariance). These parameters compound an intermediate subspace $\mathbf{\theta_{c}}$ and can be expressed by linear combinations of a set of basis vectors $\mathbf{w_{\mu}}$ or $\mathbf{w_{\sigma}}$ guided by semantic attributes ($\mathbf{a_{c}}$). For example, $\mathbf{\mu_{c}} = f_{\mu}(\mathbf{a_{c}}) = \mathbf{W_{\mu}a_{c}}$. The vector basis $\mathbf{W_{\mu}} = [ \mathbf{w_{\mu_{1}}}, \mathbf{w_{\mu_{2}}},..., \mathbf{w_{\mu_{K}}}]$, are learned from attributes or word vectors and the empirical estimates of $\mathbf{\hat{\mu}}$ and $\mathbf{\hat{\sigma}^{2}}$ are acquired directly from data with \gls{mle} or \gls{map} using linear models (e.g., least square regression) or non-linear model (e.g., kernel regression). These basis vectors can be learned only from seen classes and exploited to unseen action classes. More recently, \citet{mishra:2020-ref73} investigated the zero-shot learning recognition problem using synthesized features with two distinct approaches: \gls{iaf} and bi-directional \gls{gan}. The key idea of the approaches is to generate latent features from attributes or word vectors and then perform zero-shot learning into this embedding space in a supervised manner.

A different strategy for creating a common visual-semantic intermediate space is introduced in~\citep{qin:2017-ref35}. It is based on \gls{ecoc} specifically designed for zero shot learning (ZSECOC). These codes are learned from seen classes by latent factor decomposition and joint optimization\footnote{We recommend consulting the original paper for mathematical details.}. The codes are represented as $\mathbf{B}$ with $B = \{b_{i}\}_{i=1}^{|\mathcal{S}|} \in \{-1, 1\}^{m \times |\mathcal{S}|}$, where $m$ is the dimension of the codes and $|\mathcal{S}|$ is the number of seen classes. Aiming to relate seen $\mathcal{S}$ and unseen classes $\mathcal{U}$, the semantic similarity $s_{i,j}$ between each pair of classes in word embedding space is computed with cosine distance in order to transfer these relationships to the codes. This information is stored in a similarity matrix $\mathbf{S}$. Hence, we have $\mathbf{S^{u}} = \{ s_{i,j}^{u} \} \in \mathbb{R}^{\mathcal{S} \times \mathcal{U}}$ and $\mathbf{B^{u}}$ is given by $B^{u} = sign(BS^{u})$ where $sign(\cdot)$ is 1 if the argument is positive and 0 otherwise. The classification is performed by learning binary classifiers based on all the seen data $\mathbf{x}$ and the associated labels. This result in $m$ independent classifiers, one to each bit in the codes $F(\mathbf{x^{u}})$. The assignment of a label to unseen instance is done as
\begin{equation}
    y^{*} = argmin_{j} d_{H}(F(x^{u}), b_{j}^{u})
    \label{eq:zsecoc}
\end{equation}
\noindent where $d_{H}$ denotes the Hamming distance between prior codes and new codes (i.e., predicted with the classifiers).

\textcolor{nred}{Many works addressed the relationship between actions and objects or scenes with relative success.} For example, \citet{guadarrama:2013-ref46} proposed an approach based on hierarchical semantic models, where hierarchies are learned to subjects, objects and verbs. Thus, the training step consists of associating visual information with the corresponding leaf in the hierarchy. More specifically, the \gls{dtf} method is performed to extract handcrafted features and learn a codebook for the entire video. Object detectors from~\citep{felzenszwalb:2010} and~\citep{li:2010} are used to select the maximum score assigned to each object in any frame. A multi-channel approach combines activities and descriptors of subjects or objects sending this information to a non-linear \gls{svm}. Once the leaf classifiers are trained, the nodes are predicted by trading off specificity with semantic similarity, evaluating how semantically close the predicted triplet is to the true action. Therefore, the posterior probabilities of internal nodes are obtained by learning one-vs-all \gls{svm} classifiers for the leaf nodes and summing them. With these values, the WUP similarity (from \citet{wu:1994} work) is computed and the better triplet is predicted.

\citet{jain:2015-ref31} also described the actions by calculating the detection probability of objects in the video frames utilizing a \gls{cnn} trained with the Imagenet Dataset. In their method, an entire dataset can be classified without prior knowledge of any action class. Therefore, each video $v$ is represented as $p_{v} = [ p(y_{o_{1}}, v), \ldots, p(y_{o_{m}},v)]^{T}$, where $y_{o_{m}}$ denotes the $m\textit{-th}$ object class and $p(y_{o_{m}},v)$ is the average of the frame object  probabilities\footnote{They utilized 15,293 object categories.} (i.e., the output of the \gls{cnn}) sampled every $10$ frames. The affinity between an object class $y_{o}$ and an action class $y$ is given by $g_{y_{o}y} = s(y_{o})^{T}s(y)$, where $s(\cdot)$ is a semantic embedding of any class (i.e., object class or action class) from Word2Vec. The semantic description of an action class $y$ in function of the object classes is $g_{y} = [s(y_{o_{1}}), \ldots, s(y_{o_{m}})]^{T}s(y)$ and the vector representation of the $k$ most related objects can be estimated by Fisher Vectors. The classification consists of sampling spatio-temporal segments in a video, $U_{st}$, and applying the following
\begin{equation}
    C(v) = \argmax_{y \in Y, u \in U_{st_{v}}}\sum_{y_{o}}p_{uy_{o}}g_{y_{o}y}.
    \label{eq:jain-1}
\end{equation}

\citet{zuxuan:2016-ref71} proposed a simple but effective approach to generating an intermediate space that represents the relationships among objects, scenes and actions. In their method, a semantic fusion network fuse three streams: global low-level \gls{cnn} (e.g., from a VGG19 trained on ImageNet); 
object features in frames (e.g., from a VGG19 trained on a subset of 20574 objects); and features of scenes (e.g., from a VGG16 trained on Places 205 dataset). These three features are extracted at the frame level, and an average operation computes the scores for the videos. After, the joint features are used to train a three dense layer network composed of two hidden layers and one softmax layer. The correlation between objects/scenes and video classes is mined from the visualization of the network by salience maps. This procedure produces a matrix with the probability of each pair (object, scene) is related to an action.

\citet{mettes:2017-ref62}, on the other hand, proposed a method to classify actions without any video example in the training phase. The method is based on spatial-aware object embeddings, i.e., action tubes scored from interactions between actors and local objects. As prior knowledge, they utilize actions, actors, objects, and their interactions. The similarity between object classes and action classes is provided by the cosine distance from their Word2Vec representations. They proposed a new representation between actor and object exploring where objects tend to occur relative to the actor. This information is acquired using the MS-COCO dataset and the Faster R-CNN for detection of both objects and actors. 
They also proposed ways to scoring bounding boxes with object interaction and to link spatial-aware boxes into video tubes (i.e., bounding boxes that localize the actions and their related objects in the space and time). To distinguish tubes from different videos, they utilized global object classifiers through the GoogleLeNet network. The predicted class for a video sample is determined as the class with the highest combined score (i.e., video tube embeddings and global classifiers).

\citet{gao:2019-ref70} introduced a new strategy to model the semantic relationship between action-attribute\footnote{In their work, objects are considered attributes.}, action-action, attribute-attribute. \gls{gcn} \citep{kipf:2017} are used in a two-stream configuration. The first stream is responsible for learning classifiers on graph models constructed with ConceptNet5.5~\citep{speer:2017} and where the concepts are represented with word vectors in order to have a fixed-length representation. \textcolor{nred}{In the second stream, the visual representations of objects (acquired with a combination of the methods used by~\citet{jain:2015-ref31} and~\citet{mettes:2017-ref62}) are employed to construct graphs.} During training, the classifiers are optimized for the seen categories and are also generalized to zero-shot categories via relationship modeling. At the testing phase, the generated classifiers of unseen categories (i.e., from the first stream) are used to perform the classification on the object features of test videos (i.e., from the second stream).

\textcolor{nblue}{By observing that ConceptNet has no representations to phrase labels (e.g., playing guitar), \citet{ghosh:2020-ref74} proposed a novel method to learn knowledge graphs applied to actions. In their method, knowledge graphs are fed to a \gls{gcn}, and the training objective consists in to minimize the distance between the final classifier layer weights from \gls{gcn} with the classifier weights layer from \gls{i3d}. The adopted metric was \gls{mse}.}

\textcolor{nblue}{\citet{jones:2019-ref76} proposed a method to generate semantic embedding spaces based on dynamic attributes signatures. Their method assumes that static attributes are not suitable for modeling actions due to the lack of temporal information. Therefore, finite state machines were constructed over the static annotations provided in the UCF101 and Olympic Sports datasets. For example, they modeled five possible states for each provided attribute: (0): Absence, (1): Persistence, (2): Start, (3): End, and (4): Sometimes. Each state machine contains the transition rules and corresponds to action signatures. The classification is performed by a sequence-level score function over a pre-defined $M$ hypothesized segments. The authors show that this method can also be applied to classification and segmentation tasks.}

\textcolor{nred}{Finally, some methods explore multi-modal learning by using video and text pairing.} In~\citep{zhang:2018-ref28}, hierarchical sequential data from videos and text descriptions are modeled. 
The authors extended the general flat sequence embedding approach that is extensively used (e.g., in video understanding or video captioning). In the original model, paragraphs (i.e., a set of sentences) were represented as a sequence of words that are used in an encoder (e.g., \gls{lstm} units or \gls{gru}) to obtain a paragraph embedding. Similarly, videos are a sequence of short clips composed of frames that are used by an encoder to obtain a video embedding.

In this general scheme, the global alignments between the representations are evaluated with a loss function at a high level (e.g., cosine distance). The extension proposed is to add a mid-layer between paragraphs and their embeddings, and between videos and their embeddings. The paragraphs are encoded as a sequence of sentences and the sentences as words (i.e., there are two encoders). In addition to global alignment, local alignments are calculated for mid-layers. The quality of the intermediate encoding is improved by using decoding networks to evaluate reconstruction errors. The \gls{zsl} classification occurs through video encoding functions over visual data and textual information alignment.

\citet{piergiovanni:2018-ref57} also developed a method to learn an intermediate representation for both videos and texts based on an encoder-decoder approach. In their method, there are two pairs of encoder-decoders: (video-encoder) $E_{v}: v \xrightarrow{} z_{v}$ and (video-decoder) $G_{v}: z \xrightarrow{} v$; and (text-encoder) $E_{t}: t \xrightarrow{} z_{t}$ and (text-decoder) $G_{t}: z \xrightarrow{} t$. They used four loss functions (reconstruction, joint, cross-domain, and cycle\footnote{We suggest reading the original paper for more details.}) to properly treat the learning with paired and unpaired data. The data is paired if we have a pair of video and their descriptive sentence and it is unpaired otherwise. The unpaired learning is conducted in a semi-supervised manner based on adversarial learning by defining three networks (i) to discriminate between text and video-latent representations, (ii) to discriminate the generated video data from the textual information, and (iii) to discriminate the generated textual data from visual information. \textcolor{nred}{This last discriminator is especially important when the testing is conducted in datasets such as the UCF101 and HMDB51 that have no captions available because it enables the knowledge transfer.} The ActivityNet Captions and Charades Datasets provided the sentences used in the learning process. Once the model is learned, \gls{zsl} classification is conducted by the nearest neighbour rule between each video representation $z_{v}$ and its text representation $z_{t}$ in the intermediate space.

\subsection{Classification into the visual embedding space}

\textcolor{nred}{We identify that some recent methods attempt to synthesize the visual features for unseen classes using the features of seen ones and the semantic information. These approaches differ from the two priors because the function learned uses the visual information and the semantic information in the reverse direction. That is, instead of projecting onto semantic space or an intermediate space, the output is given in the visual domain, taking advantage of conditional adversarial learning.} For example, \citet{zhang:2018-ref65} proposed a multi-level semantic inference method to tackle the problem of modeling the joint distribution of visual features and semantic knowledge and a matching-aware mutual information correlation to solve the semantic gap by transferring semantic knowledge. Briefly, a group of noise is used to synthesize video features, which is simultaneously used by the inference model and the discriminator $D$ to perform semantic inference and correlation constraint. This inference model is responsible for learning an inverse mapping from the synthetic video feature to corresponding semantic knowledge. In the discriminator, two embeddings (i.e., matched and mismatched) are evaluated. After the adversarial training, the model produces visual features classified with the nearest neighbor by evaluating the distance between the generated output and the original visual feature. It is possible to use \gls{svm}s and, in this case, the visual features of unseen (i.e., synthesized) and seen categories are merged to train the model in a supervised manner.

\citet{mandal:2019-ref72} investigated the case of \gls{fsl} and addressed \gls{zsl} as a special case. 
They proposed to classify if an instance came from the seen or unseen dataset using an out-of-distribution classifier to produce a non-uniform distribution with emphasis on seen categories and a uniformly distributed output on the seen categories. However, to zero-shot learning case, this out-of-distribution classifier is not used and their method became similar to~\citep{zhang:2018-ref65}, but adapting a Wasserstein \gls{gan} \citep{arjovsky:2017} conditioned on the embeddings of seen class labels (i.e., in training) and unseen (i.e., in testing). The \gls{zsl} classification is made by a classifier that maps the synthesized features to the unseen class labels.

\section{Benchmark Datasets}
\label{sec:datasetsprotocol}

\begin{table*}
    \centering
    \caption{\textcolor{nblue}{Datasets used in the ZSAR experiments ordered by year of creation. The number of videos (\#V) and the number of classes (\#C) are also provided for each dataset.}}
    \footnotesize
    \color{nblue}
    \resizebox{1.0\textwidth}{!}{
    \begin{tabular}{p{2.5cm}p{0.8cm}p{0.8cm}p{0.5cm}p{9.6cm}}
		\hline
		Datasets                   &    Year   & \#V  & \#C  & Used in papers \\ 
		\hline
		
		KTH~\citep{schuldt:2004}               & 2004   & 2391     &  6        &~\citet{liu:2011-ref58} \\
		Weizmann~\citep{blank:2005}            & 2005   & 81       &  9        &~\citet{liu:2011-ref58,qiu:2011-ref11}\\
		UCFSports ~\citep{rodriguez:2008}      & 2008   & 150      & 10        &~\citet{qiu:2011-ref11,jain:2015-ref31}\\
		UIUC~\citep{tran:2008}                 & 2008   & 532      & 14        &~\citet{liu:2011-ref58} \\
		Olympic Sports~\citep{niebles:2010}    & 2010   & 800      & 16        &~\citet{liu:2011-ref58,xu:2016-ref23,xu:2017-ref24,qin:2017-ref35,mishra:2018-ref25,gao:2019-ref70,mandal:2019-ref72,mishra:2020-ref73,jones:2019-ref76} \\
		UCF50~\citep{reddy:2013}               & 2010   & 6676     & 50        &~\citet{qiu:2011-ref11} \\
		CCV~\citep{jiang:2011}                 & 2011   & 9317     & 20        &~\citet{xu:2017-ref20,xu:2017-ref24} \\
		HMDB51~\citep{kuehne:2011}             & 2011   & 7000     &    51     &~\citet{xu:2015-ref19,xu:2016-ref23,xu:2017-ref24,jain:2015-ref31,alexiou:2016-ref54,wang:2017-ref14,wang:2017-ref18,qin:2017-ref35,mishra:2018-ref25,piergiovanni:2018-ref57,zhu:2018-ref55,roitberg:2018-ref56,hahn:2019-ref43,gao:2019-ref70,bishay:2019-ref78,mandal:2019-ref72,mishra:2020-ref73,ghosh:2020-ref74,brattoli:2020-ref75}\\
		UCF101~\citep{soomro:2012}             & 2012   & 13320    &    101    &~\citet{xu:2015-ref19,kodirov:2015-ref22,gan2015-ref21,xu:2017-ref20,jain:2015-ref31,xu:2016-ref23,alexiou:2016-ref54,xu:2017-ref24,wang:2017-ref14,wang:2017-ref18,qin:2017-ref35,mishra:2018-ref25,piergiovanni:2018-ref57,zhu:2018-ref55,roitberg:2018-ref56,hahn:2019-ref43, gao:2019-ref70,bishay:2019-ref78,mandal:2019-ref72, mishra:2020-ref73,jones:2019-ref76,ghosh:2020-ref74,brattoli:2020-ref75}\\
		MPII CC~\citep{rohrbach:2012-ref12, rohrbach:2013-ref39} & 2012 & 256 & 41  &~\citet{rohrbach:2012-ref12,rohrbach:2013-ref39} \\
		Thumos14~\citep{idrees:2017}           & 2014   & 1574     & 101       &~\citet{jain:2015-ref31}  \\
		Breakfast ~\citep{kuehne:2014}         & 2014   & 1989     & 10        &~\citet{wang:2020-ref15}  \\
		ActivityNet~\citep{heilbron:2015}      & 2015   & 27801    &   203     &~\citet{zhang:2018-ref28,piergiovanni:2018-ref57, zuxuan:2016-ref71} \\
		Charades ~\citep{sigurdsson:2016}      & 2016   & 9848     & 157       &~\citet{wang:2020-ref15,ghosh:2020-ref74}  \\
		Kinetics 400~\citep{kay:2017}          & 2017   & 306245   & 400       &~\citet{hahn:2019-ref43} \\
		MLB-YouTube~\citep{piergiovanni:2018}  & 2018   & 4290     & 8         &~\citet{piergiovanni:2018-ref57} \\
		Kinetics 700~\citep{carreira:2019}     & 2019   & 650000   & 700       &~\citet{brattoli:2020-ref75} \\ 
		\hline
	\end{tabular}
	}
	\label{tbl:datasetsoverview}
\end{table*}

The first popular video benchmarks were small, with approximately 10k videos~\citep{carreira:2017}, as shown in Table~\ref{tbl:datasetsoverview}. Larger and complex datasets are available since 2011, such as HMDB51~\citep{kuehne:2011}, UCF101~\citep{soomro:2012}, ActivityNet~\citep{heilbron:2015} and, more recently, Kinetics~\citep{kay:2017,carreira:2017}.

KTH~\citep{schuldt:2004} is a dataset with six types of human actions (walking, jogging, running, boxing, hand waving and hand clapping) performed by 25 different people in four different scenarios (outdoors, outdoors with scale variation, outdoors with different clothes, and indoors). The dataset contains 2,391 sequences taken over homogeneous backgrounds with a static camera and a frame rate of 25 \gls{fps}. \textcolor{nred}{This dataset is no longer challenging and has not been used to evaluate modern \gls{zsar} methods}. Another simple dataset is the Weizmann~\citep{blank:2005} with nine types of actions (running, walking, jumping-jack, jumping-forward-on-two-legs, jumping-in-place-on-two-legs, galloping-sideways, waving-two-hands, waving-one-hand, and bending) performed by nine different people in low-resolution videos (180$\times$155) with 25 \gls{fps}.

\textcolor{nred}{KTH and Weizmann datasets contain a single staged actor with no occlusion and low clutter. They present video clips with controlled illumination and camera position so that they are not quite representative of the complexity of the real-world scenario and are not used recently. To address these limitations, \citet{kuehne:2011} presented the HMDB51 dataset with videos from many sources such as digitized movies, Prelinger archive, YouTube, and Google videos}. This dataset contains 51 actions grouped into 5 categories (general facial actions, facial actions with object manipulation, general body movements, body movements with object interaction, and body movements for human interaction). The height of all the frames is scaled to 240 pixels, and so the width is rescaled, keeping the original aspect ratio. The frame rate is converted to 30 \gls{fps} in order to ensure consistency in the entire dataset. \textcolor{nred}{Due to the complexity of the videos and significant number of videos per class, this dataset is widely used for evaluation}.

There are three datasets provided by the \gls{ucf} that are used in \gls{zsar}: UCFSports~\citep{rodriguez:2008}, UCF50~\citep{reddy:2013} and UCF101~\citep{soomro:2012}. 
In these datasets, the complexity grows because the videos are taken from the Web and they contain random camera motion, poor lighting conditions, clutter, as well as changes in scale, appearance and viewpoints, and occasionally no focus on the actions of interest~\citep{reddy:2013}. UCFSports, for example, contains 10 actions (diving, golf swing, kicking, lifting, riding horse, running, skateboarding, swing-bench, swing-side and walking) distributed in 150 video sequences with a resolution of 720$\times$480 and 10 \gls{fps}. This dataset was collected from various sports featured on broadcast television channels, such as BBC and ESPN. On the other hand, UCF50, an extension of the UCF11 dataset~\citep{liu:2009}, contains 50 categories with a minimum of 100 videos for each action class and a total of 6,676. Finally, UCF101~\citep{soomro:2012} has 101 action classes with a total of 13,320 videos with frame resolution standardized to 25 \gls{fps} and resolution to 320$\times$240 pixels and stored in \texttt{avi} format. The action categories are divided into five types (human-object interaction, body-motion only, human-human interaction, playing musical instruments, and sports) and grouped into 25 groups where each group consists of 4-7 videos of an action. \textcolor{nred}{This great variation of action types and the largest amount of examples make this dataset widely used in experiments, as well as HMDB51.}

Olympic Sports~\citep{niebles:2010} is a complex dataset of activities collected from YouTube sequences. There are 16 activities with 50 sequences per class, and the complex motions go beyond simple punctual or repetitive actions in contrast to UCFSports~\citep{rodriguez:2008}, which contains periodic or simple actions such as walking, running, golf-swing or ball-kicking. Although proposed for activity recognition, this dataset was used in approaches that focus on action recognition~\citep{liu:2011-ref58, xu:2017-ref20, xu:2016-ref23, xu:2017-ref24, qin:2017-ref35, mishra:2018-ref25}, \textcolor{nred}{demonstrating that the complexity of methods makes them able to work on simple activities.} \gls{ccv} is a dataset introduced by~\citet{jiang:2011} and includes 9317 unconstrained videos from the Web, preserving the originality without post-editing. There are 20 semantic categories, including a broader set ranging from events, objects, to scenes annotated using the \gls{amt} platform. The number of videos from each category varies from 200 to 800. \textcolor{nred}{This dataset is used only in few works~\citep{xu:2017-ref20, xu:2017-ref24} because there are examples of actions, activities, objects, and events, being more indicated to video description or retrieval problems. Another limitation is the few number of actions}. For example, if a standard protocol that divides in 50\% as seen and the rest of unseen classes are performed, the result is a restrict visual space and poor global performance.

MPII Cooking Composites and Breakfast are datasets that contain only cooking activities. MPII Cooking Composites contains 41 basic cooking activities with varying length from 1 to 41 minutes distributed on 256 videos. \textcolor{nred}{However, this dataset was used only in the same work where it was introduced.} Likewise, Breakfast~\citep{kuehne:2014} is a large dataset of daily cooking activities, including a total of 52 participants performing 10 activities in 18 real-life kitchens. The resolution is 320$\times$240 pixels with 15 \gls{fps}. \textcolor{nred}{This dataset was used in a work that explores multi-label zero-shot action recognition~\citep{wang:2020-ref15} because there are 49 action classes annotated\footnote{Actions that compound the ten cooking activities.} in the clips and more than one action per clip}. Charades dataset~\citep{sigurdsson:2016} is also used in~\citep{wang:2020-ref15} and has activities composed of more than one action. Charades is a challenging dataset built with the collaboration of 267 persons from three continents by using the \gls{amt} platform. The objective was to collect videos of common daily activities performed in their homes -- especially, examples that are not easy to find on YouTube, movies, or TV broadcasts. The dataset has 9,848 annotated videos representing 157 actions with 30 seconds of duration each. \textcolor{nred}{However, as most works do not explore multi-label classification, these datasets are not used for evaluation}.

The ActivityNet dataset was introduced by~\citet{heilbron:2015} and is a large-scale benchmark for human activity understanding. There is a range of complex human activities that are of interest to people in their daily living. More precisely, 203 activity classes with an average of 137 untrimmed videos per class and a total of 27,801 videos. These videos were collected from the Internet, exploring a large amount of video data on online repositories such as YouTube. Around 50\% of the videos have a resolution of 1280$\times$720, whereas the majority has 30 \gls{fps}. \textcolor{nred}{This dataset is little explored, possibly due to its high complexity compared to their amount of videos per class (193 on average)}. It is used in recent works~\citep{zhang:2018-ref28,piergiovanni:2018-ref57,zuxuan:2016-ref71}, which explore multi-modal learning by combining visual features with textual descriptions. On the Kinetics dataset~\citep{kay:2017}, which is the most extensive collection of human actions available to benchmark, there are 400 complex human action classes from different YouTube videos with at least 400 video clips for each action. The clips are about 10 seconds long, variable resolutions and frame rates. \textcolor{nred}{This dataset can be considered the successor of HMDB51, UCF101, and ActivityNet (trimmed version) because it is more suitable for training deep networks from scratch}. The HMDB51 and UCF101 datasets are not large enough or have sufficient variation to learn and evaluate the current generation of human action classification models based on deep learning, and this limitation is more evident in zero-shot action recognition. More recently, an extension called Kinetics 700~\citep{carreira:2019} was used in~\citep{brattoli:2020-ref75}.

As shown in Table~\ref{tbl:datasetsoverview}, there is a group of datasets used only once, such as UIUC~\citep{tran:2008}, Thumos14~\citep{idrees:2017}, and MLB-YouTube~\citep{piergiovanni:2018}. 
The UIUC dataset is presented in~\citep{tran:2008}. It consists of 532 high-resolution sequences of 14 activities performed by 8 actors in a single view. The Thumos14 dataset was proposed in the Thumos Challenge context \citep{thumos:2014}. In this dataset, there are temporally untrimmed videos and background videos, that is, with a similar background but without actions in the scene. The 101 action classes are performed in realistic settings and distributed in 1574 video clips. MLB-YouTube~\citep{piergiovanni:2018} is a dataset with activities collected from broadcast baseball videos with a focus on fine-grained activity recognition. More precisely, it is composed of 20 baseball games (42 hours) from the 2017 MLB post-season available on YouTube. In this dataset, the structure of the scene is very similar among activities; often, the only difference is the motion of a single person. Additionally, there is a single camera viewpoint to determine the activity. Due to its objective, this dataset has limited potential in \gls{zsl}. A complete description of most of datasets can be found in~\citep{chaquet:2013,kang:2016,singh:2019}.

\section{Experimental Protocols and Performance Analysis}
\label{sec:performance}

There are many experimental protocols to perform \gls{zsar} in videos. Consequently, it is not easy to compare them. We select works that use HMBD51, UCF101 or Olympic Sports datasets since they are the most popular, as shown in Table~\ref{tbl:datasetsoverview}, which enables comparison among different approaches. The ActivityNet was not selected because we do not discriminate between its two versions (i.e., trimmed and untrimmed) in Table~\ref{tbl:datasetsoverview}. \textcolor{nblue}{Table~\ref{tbl:performance} reports the results of selected works, the proportions and amount of runs used in experiments and a comparison of performance in inductive and transductive settings. Additionally, we provide complementary information on how the visual and semantic embedding were performed to acquire that result and what was the classification approach.}

The approaches are commonly evaluated using a general strategy. Initially, the classes of the dataset are randomly split into two disjoint sets called seen (source) and unseen (target) with different proportions (90\%/10\%, 80\%/20\% and 50\%/50\%).
This procedure is repeated many times (3, 5, 10, 30, 50) and, in none work, the chosen proportions and/or the number of runs are justified. \textcolor{nblue}{We identify three performance metrics reported (i.e., overall accuracy, mean per-class accuracy, and mean average precision). We included the value, rounded to one decimal place, the standard deviation, when was reported, and the measure type.}

The motivations for the use of 90\%/10\%, 80\%/20\% or 50\%/50\% splits in each dataset is not clear. It is reasonable to think in terms of the size of the training split. That is, to evaluate whether the method presents better results in the presence of more training information. However, at the same time that they use more information to learn, there are fewer examples to classify and the results tend to be better. On the other hand, in a configuration of 50\%/50\%, the results tend to be worse because there are more examples to classify and less information available to learn the models. This behavior is clearly identified in Table~\ref{tbl:performance}. In large scale datasets, such as HMDB51 or UCF101, with 50\%/50\% configuration, it is possible to obtain a relevant amount of videos for both to learn and classify. Thus, this configuration is widely used. \textcolor{nblue}{Due to the domain shift problem, few works have adopted cross dataset configurations, where the model is trained in one dataset and is evaluated in another. An example is shown in Table~\ref{tbl:performance} marked as 0/20 and 0/50 with impressive results compared with intra-class approaches. Their work performs transfer learning by leveraging object-action relationships.}

\begin{sidewaystable*}
%\begin{table}
	\caption{\textcolor{nblue}{ZSAR performance on the HMDB51, UCF101 and Olympic Sports datasets.The results are presented rounded to one decimal place for both mean value ($\bar x$) and standard deviation ($s$), when this value is presented in the original paper. Visual embedding (VE); Semantic embeding (SE); Classification strategy (C); Inductive setting (I); Transductive setting (T); Improved dense trajectories (IDT); Convolutional 3D network (C3D); Inflated 3D network (I3D); Object detector (OD); Attributes (A); Word2Vec (W2V); Global vectors (GloVe); Sentence to vector (S2V); Fisher feature vector (FFV); Classification into the semantic space (SS); Classification into an intermediate space (IS); Classification into the visual space (VS); Overall accuracy (Acc.); Mean per-class accuracy (P-c Acc.), and Average precision (AP). $^{*}$ indicates that, in this experiment, is not possible to estimate $\mu = \bar x \pm 1.0$ with 95\% of confidence. $^{\dagger}$ indicates that, in this experiment, is not possible to estimate $\mu = \bar x \pm 2.0$ with 95\% of confidence.}}
	\label{tbl:performance}
	\centering
	\small
	\color{nblue}
	\begin{tabular}{p{0.5cm}p{0.4cm}p{1.1cm}p{0.9cm}p{0.4cm}p{1.2cm}p{1.5cm}p{1.5cm}p{1.5cm}p{1.5cm}p{1.5cm}p{1.5cm}p{2.9cm}}\hline
		\%     & \#   &  VE  & SE    & C  & Metric     &  \multicolumn{2}{c}{HMDB51}          & \multicolumn{2}{c}{UCF101}            & \multicolumn{2}{c}{Olympic Sports} & Reference                       \\ 
		       &      &      &       &    &            &    I              &      T           &     I             &           T       &     I           &    T             &                                \\
		\hline
		
		90/10  & 3    & C3D  & W2V   & SS & Acc.       & $51.9$            & $-$              & $49.4$            & $-$               & $-$             & $-$              & \citet{hahn:2019-ref43}        \\ 
		       & 5    & IDT  & W2V   & IS & AP         & $-$               & $-$              & $81.8$            & $-$               & $-$             & $-$              & \citet{gan2015-ref21}          \\
		       &      & I3D  & S2V   & IS & Acc.       & $-$               & $-$              & $69.6$            & $-$               & $-$             & $-$              & \citet{ghosh:2020-ref74}       \\
		\hline
		
		80/20  & 3    & C3D  & W2V   & SS & Acc.       & $38.2$            & $-$              & $37.4$            & $-$               & $-$             & $-$              & \citet{hahn:2019-ref43}        \\ 
		 	   & 10   & IDT  & A+W2V & SS & Acc.       & $-$               & $-$              & $22.5\pm3.5^{\dagger}$ & $-$          & $-$             & $-$              & \citet{kodirov:2015-ref22}     \\
		  0/20 &      & OD   & W2V  & IS  & Acc.       & $-$               & $-$              & $51.2\pm5.0^{\dagger}$ & $-$          & $-$             & $-$              & \citet{mettes:2017-ref62}      \\
		       & 30   & C3D+IDT & A+W2V & SS & P-c Acc. & $-$               & $-$              & $51.1\pm1.2$ & $66.9\pm1.9$ & $-$   & $-$        & \citet{wang:2017-ref18}        \\
		       &      & C3D  & A     & SS  & Acc.      & $-$               & $-$              & $42.7\pm5.4^{\dagger}$ & $-$          & $-$             & $-$              & \citet{bishay:2019-ref78}      \\
     	\hline
		
		50/50  & 3    & C3D  & W2V   & SS & Acc.        & $24.1$            & $-$              & $22.0$            & $-$               & $-$             & $-$              & \citet{hahn:2019-ref43}        \\
		       & 5    & IDT  & W2V   & IS & Acc.       &$19.7\pm1.6^{*}$&$24.8\pm2.2^{\dagger}$&$18.3\pm1.7^{\dagger}$&$22.9\pm3.3^{\dagger}$& $-$ & $-$ &\citet{xu:2016-ref23} \\
		       & 5    & IDT  & W2V   & IS & AP         & $-$               &  $-$              &  $-$              & $-$  &$44.3\pm8.1^{\dagger}$&$56.6\pm7.7^{\dagger}$    &\citet{xu:2016-ref23} \\
      	       &      & C3D  & FFV   & IS & Acc.       & $25.8\pm1.2^{*}$  & $31.5\pm1.7^{\dagger}$ & $40.1\pm1.3^{*}$ & $50.6\pm2.5^{\dagger}$ & $-$   & $-$              & \citet{wang:2017-ref14}        \\
 		       & 10   & IDT  & W2V   & SS  & Acc.      & $14.4$            & $22.4 $          & $12.0$            & $35.2$            & $-$             & $-$              & \citet{alexiou:2016-ref54}     \\
		  0/50 &      & OD & W2V & IS   & Acc.         & $-$ & $-$              & $40.4\pm1.0$      & $-$               & $-$             & $-$              & \citet{mettes:2017-ref62}      \\
		       &      & IDT  & A+W2V & SS & Acc.       & $-$    & $-$              & $14.0\pm1.8^{*}$  & $-$               & $-$             & $-$              & \citet{kodirov:2015-ref22}     \\
		       
		       & 30   & DTF  & W2V   & SS & Acc.       & $18.0\pm3.0^{*}$  & $21.2\pm3.0^{*}$ & $12.7\pm1.6$      & $18.6\pm2.2$      & $-$             & $-$              & \citet{xu:2015-ref19}          \\
	           &      & C3D  & A     & IS & Acc.       & $-$            & $-$              & $22.7\pm1.2$      & $24.5\pm2.9^{*}$  &$50.4\pm11.2^{\dagger}$&$57.9\pm14.1^{\dagger}$& \citet{mishra:2018-ref25}\\
	           &      & C3D  & W2V   & IS & Acc.       & $19.3\pm2.1$   & $20.7\pm3.1^{*}$ & $17.3\pm1.1$      & $20.3\pm1.9$     & $34.1\pm10.1^{\dagger}$&$41.3\pm11.4^{\dagger}$& \citet{mishra:2018-ref25}\\
		       &      & C3D+IDT & A+W2V & SS & P-c Acc. & $-$             & $-$              & $26.4\pm0.6$      & $35.1\pm 1.1$ & $-$    & $-$              & \citet{wang:2017-ref18}        \\
		       &      & C3D+IDT & W2V & SS & P-c Acc.  & $20.6\pm0.8$      & $22.3\pm1.1$     & $-$               & $-$               & $-$             & $-$              & \citet{wang:2017-ref18}        \\
		       &      & C3D  & A     & SS & Acc.       & $-$               & $-$              & $23.2\pm2.9^{*}$  & $-$               & $-$             & $-$              & \citet{bishay:2019-ref78}      \\
		       &      & C3D  & W2V   & SS & Acc.       & $19.5\pm4.2^{*}$  & $-$              & $19.0\pm2.3$      & $-$               & $-$             & $-$              & \citet{bishay:2019-ref78}      \\
               &      & I3D  & A     & VS & P-c Acc.   & $-$               & $-$              & $38.3\pm3.0^{*}$  & $-$               & $65.9\pm8.1^{\dagger}$  & $-$      & \citet{mandal:2019-ref72}      \\
               &      & I3D  & W2V   & VS & P-c Acc.   & $30.2\pm2.7^{*}$  & $-$              & $26.9\pm2.8^{*}$  & $-$               & $50.5\pm6.9^{\dagger}$  & $-$      & \citet{mandal:2019-ref72}      \\
		       
		       &      & C3D  & A     & IS & Acc.       & $-$ & $-$         &$25.2\pm3.0^{*}$ & $26.1\pm3.0^{*}$&$52.1\pm11.7^{\dagger}$&$54.9\pm11.7^{\dagger}$&\citet{mishra:2020-ref73} \\
   	           &      & C3D  & W     & IS & Acc.       & $17.5\pm2.4$    &$21.3\pm3.2$    & $-$               & $-$               & $-$             & $-$              &\citet{mishra:2020-ref73} \\
		       &      & OD   & A     & IS & P-c Acc.   & $-$&$-$&$48.9\pm5.8^{\dagger}$ &$48.9\pm5.8^{\dagger}$&$74.2\pm9.9^{\dagger}$&$74.2\pm9.9^{\dagger}$& \citet{jones:2019-ref76} \\

		       &      & IDT  & A+W   & SS & Acc.       & $-$ &$-$& $11.7\pm1.7$ & $22.1\pm2.5$ & $51.7\pm11.3^{\dagger}$ &$53.2\pm11.6^{\dagger}$         & \citet{xu:2017-ref24}          \\
		 	   &      & IDT  & W     & SS & Acc.       & $14.5\pm2.7$      & $ 24.1\pm 3.8^{*}$& $-$              & $-$               & $-$             & $-$              & \citet{xu:2017-ref24}          \\
		 	   
		 	   &      & IDT  & GloVe & VS & Acc.       & $25.3\pm4.5^{*}$  & $-$              & $25.4\pm3.1$      & $-$               & $43.9\pm7.9^{\dagger}$    & $-$    & \citet{zhang:2018-ref65}       \\
		 	   &      & VGG19&GloVe&VS & Acc.          & $21.6\pm5.5^{*}$  & $-$              & $28.8\pm5.7^{*}$  & $-$               & $35.5\pm8.9^{\dagger}$    & $-$    & \citet{zhang:2018-ref65}       \\
		       &      & OD   & W2V   & IS & Acc.       & $23.2\pm3.0$&$31.0\pm3.2$&$34.2\pm3.1$&$41.6\pm3.7$&$56.5\pm6.6^{*}$&$59.9\pm5.3^{*}$& \citet{gao:2019-ref70} \\
		\hline
	\end{tabular}
%\end{table}
\end{sidewaystable*}

%%% \citet{ghosh:2020-ref74}  --- ucf101 = 50.13; hmdb51 = 40.77; charades = 18.21 * protocolo especifico
%%% 50/50 sem random splits ----> ucf101 = 49.2   100/0    39.8 (obs.: traina com um valor entre 500 e 600 classes!!!) -- como reportar esses resultados?

Large scale datasets are necessary to learn more discriminative models, but the amount of all possible combinations of splits (seen/unseen) for each experiment is enormous. Therefore, it is impractical to perform experiments with all possible combinations and to use random splits is a valid strategy. In this scenario, it is necessary to consider that the experiment is stochastic and that 5 or 10 random splits can be an insufficient sampling compared to all possible combinations. For example, considering the 90/10 split, how statistically significant is a result obtained with 3 or 5 random splits? At the same time, how feasible is to perform the experiments using much more random splits?
%Hence, we only compare the results of experiments performed at least 30 times because, with this amount of runs, it is possible to provide an interval estimation for mean accuracy if it is considered a normal distribution.
Thus, we only compare the results of experiments in which the standard deviation was reported. \textcolor{nblue}{We assume that the mean accuracy has a normal distribution and approximate the population standard deviation $\sigma$ by sample standard deviation $s$, and the mean accuracy of population by $\mu \approx \bar x \pm E$, where $E \approx t_{95\%, n-1} \frac{s}{\sqrt{n}}$ and $n-1$ are the degrees of freedom for $n$ runs}. When it is impossible to estimate the mean accuracy with $1\%$ of estimation error, it is marked with $^{*}$ and, when it is impossible with $2\%$ it is marked with $^{\dagger}$.

\textcolor{nblue}{In 50/50 (seen/unseen) configuration and considering only the inductive setting, the work described by~\citet{mandal:2019-ref72} outperforms all the other methods on HMDB51. On UCF101, the works proposed by~\citet{mettes:2017-ref62} and~\citet{jones:2019-ref76} have remarkable results and are based on object-action relationships. On Olympic Sports, the works developed by~\citet{mandal:2019-ref72} and~\citet{jones:2019-ref76} show better performance. Considering the transductive setting, we highlight the results reported by~\citet{wang:2017-ref18} and~\citet{gao:2019-ref70} on HMDB51, \citet{wang:2017-ref18} and~\citet{jones:2019-ref76} on UCF101 and \citet{jones:2019-ref76} and~\citet{gao:2019-ref70} on Olympic Sports. Next, we point out some considerations on these results}.

The BiDiLEL model~\citep{wang:2017-ref18} is based on combinations of features that are projected onto an intermediate space. In the visual extraction step, \gls{c3d} deep features are combined with \gls{idt} handcrafted features and, in the semantic embedding step, a combination of attributes and Word2Vec was used on UCF101, whereas only Word2Vec was used on HMDB51, which was the most powerful combination of features available. This method was applied by~\citet{wang:2017-ref14} to explore a new semantic embedding method based on static images represented with Fisher vectors.

\textcolor{nblue}{Table~\ref{tbl:performance} shows that approaches based on simple feature representations extracted with off-the-shelf methods (i.e., Word2Vec, I3D, C3D) were outperformed by methods based on the extraction of more high-level semantic information from video clips, usually with object detection~\citep{mettes:2017-ref62,gao:2019-ref70} or multi-modal learning (i.e., combining visual information with textual descriptions~\citep{zhang:2018-ref28})\footnote{Their work has an impressive result but, due to their requirements from textual descriptions, the experiments were conducted on ActivityNet Captions dataset.}. This is a remarkable distinction between strategies adapted from object or image ZSL domain and specific strategies for action recognition in videos. Recently, several specific approaches have been proposed, leveraging ZSAR performance.}

\citet{zhang:2018-ref65} and~\citet{mandal:2019-ref72} utilized \gls{gan}s to generate more training data from the training set with the same statistic properties and to perform the classification into the visual embedding space. This strategy brings high discriminative power and suffers much less from information degradation than other methods. However, basic \gls{gan}s suffer from instability in training because they are unrestricted and uncontrollable~\citep{wang:2019-gan}. \citet{mandal:2019-ref72} adapted a Wasserstein \gls{gan} conditioned on the embeddings of seen and unseen class labels and outperformed the work described by~\citet{zhang:2018-ref65}, which demonstrates the potential of these approaches in the next years. \citet{gao:2019-ref70} explored the relationship between objects and actions using graph convolutional networks, indicating the effectiveness of using the properties from word vectors to identify relationships between objects-objects and between the objects-actions.

By evaluating the impact of transductive setting on performance, reported in Table~\ref{tbl:performance}, it is observed that this configuration presents better results than inductive setting in all works. This is due to the effectiveness of methods as self-training and hubness correction to alleviate the domain shift problem. Although exploring the manifold structure of unseen classes may improve the results, in a real world scenario, this information cannot be available and inductive approaches are preferable.

\textcolor{nblue}{Another important consideration is that the use of attributes generally results in better performance than word vectors. For example, using the same method as on UCF101, \citet{mishra:2018-ref25} obtained $22.7\pm1.2$ with attributes and $17.3\pm1.1$ with Word2Vec. \citet{jones:2019-ref76} utilized attributes, but modeling their evolution in the clips with finite state machines and acquired promising results. Nevertheless, as discussed previously, the use of attributes is not scalable and become impracticable in real-world scenarios. There is a demand for more strategies to perform semantic embedding, focusing on high-level semantic descriptions based on automatic attribute annotation, objects and scenes relationships with actions or natural language descriptions of videos.}

\section{Open Issues and Future Work}
\label{sec:open}

Although much progress has been made in zero-shot action recognition in the last years, its performance is far from conventional supervised learning. For example, while~\citet{carreira:2017} obtained 98\% and 80.9\% of accuracy on UCF101 and HMDB51 datasets using the supervised learning paradigm, respectively, \citet{hahn:2019-ref43} achieved $21.96\%$ and $24.1\%$ (50\%/50\% seem/unseen classes), respectively, using the \gls{zsl} paradigm and the same I3D model. \textcolor{nblue}{Even if we compare to the best results in \gls{zsl}, that is, those obtained by~\citet{mandal:2019-ref72} ($\sim38.3\pm1.0$) using a generative model, \citet{gao:2019-ref70} ($\sim41.6\pm1.0$) and~\citet{mettes:2017-ref62} ($\sim40.4\pm1.0$) using objects and their relationships with actions, or even~\citet{jones:2019-ref76} using dynamic attributes, we can observe that there is still a lot of room to achieve comparable or useful performance, and this requires to resolve or ameliorate the classical \gls{zsl} problem, that is, the semantic gap}.

Describing actions is much more challenging than describing nouns. Most works have explored only Word2Vec or GloVe algorithms without modifications or new techniques. \textcolor{nblue}{As shown in Figure~\ref{fig:semanticembeddingspace}, word vectors can present confusions with compound classes (e.g., pommel horse x horse riding)}. We believe that there are few variations or strategies for semantic embedding. A good example was described by~\citet{alexiou:2016-ref54}. Although the result was not globally superior than other approaches, their work demonstrated that the use of synonyms can leverage the performance of several \gls{zsl} methods. \textcolor{nblue}{Another promising  approach to consider compound labels is the sentence to vector model (Sent2Vec)~\citep{pagliardini:2018}, used in~\citep{ghosh:2020-ref74}. This model was responsible for a speedup of $\sim1.3$ compared to the results using Word2Vec in their work. Moreover, we believe that it is necessary to incorporate more recent advances in language processing, for example, geometric deep learning with Graph Convolutional Networks~\citep{yao:2017} or explore textual descriptions with transformer-based models, for instance, BERT~\citep{devlin:2019} and VideoBERT~\citep{sun:2019}.}

From the perspective of visual extraction, with the recent advances in deep learning methods, its use seems to be imperative, especially pre-trained models, recurrent networks and generative models. However, a new problem emerges. For example, the \gls{c3d} model is a pre-trained \gls{cnn} using the Sports-1M Dataset~\citep{tran:2015}. We believe that using pre-trained deep models in practice means intrinsically to use a cross-dataset approach and, if the same classes that are used to train the deep models were also used to test the \gls{zsl} methods, the disjunction between seen and unseen classes would not be respected because the deep model acquires the knowledge from classes that should be unseen. A similar analysis was presented by~\citet{roitberg:2018-ref65}, but in the context of cross-dataset studies. They argued that when external datasets are involved, one has to ensure that the terms of \gls{zsl} are still met and the seen and unseen categories are disjoint. It is not sufficient to remove only identical classes because there are similar classes, such as \texttt{Basketball Shooting} (UCF101) $\times$ \texttt{Basketball} or \texttt{Basketball 3$\times$3} or \texttt{wheelchair basketball} (Sports-1M). \textcolor{nblue}{A protocol to remove semantically similar classes from source category (seen) using the cosine similarity measure and a threshold parameter was defined by~\citet{roitberg:2018-ref65} and this analysis was extended by~\citet{brattoli:2020-ref75}}. However, when pre-trained deep models are used, it is necessary to remove the similar classes from the target and not from the source. For example, in~\citep{wang:2017-ref18}, we need to compare the classes between UCF101 and Sports-1M (used for training C3D model). It is observed that they share 23 identical classes and 17 similar classes\footnote{Manual checks.}. Since that work uses the same 30 splits employed by~\citet{xu:2015-ref19}, these shared classes were not removed from the target before the experiment, so the restriction of \gls{zsl} is not preserved. To keep the \gls{zsl} disjunction between the training and testing sets, it is necessary to use only unknown classes in the testing time, excluding all classes that were used for training the deep model. In this case, the UCF101 dataset would have 61 possible classes for testing. \textcolor{nblue}{This approach has been implemented in the work developed by~\citep{ghosh:2020-ref74}}. This new restriction means that it may be impracticable to use the UCF101 or HMDB51 dataset when pre-trained deep models, when \gls{c3d} or \gls{i3d} are used. \textcolor{nblue}{In fact, as shown in Table~\ref{tbl:datasetsoverview}, more recent datasets, such as Kinetics 600, Kinetics 700 or ARID~\citep{xu:2020arid}, have not been explored in~\gls{zsar}}.

We identify that multi-modal learning is a promising approach to address the semantic gap. However, there are few studies with this perspective. Intuitively, it is easier to recognize actions using object detection in the scene or by including more information from still images or texts because the features tend to be more descriptive, as with attributes compared to word embeddings. These alternatives need to be further explored so that we can build robust frameworks for zero-shot action recognition.

As discussed earlier, it is necessary to establish a common protocol and mainly a straight definition of the use of seen classes to fine tuning the parameters of deep models. There is a lack of works in which several experimental protocols are applied to state-of-the-art approaches, so that the community could be able to replicate and compare their results. For example, we believe that an experimental protocol is more suitable for evaluation where there is no need to randomly split the datasets. What criteria could be adopted to define which classes are used in training and which are used in testing? Is it possible to create a general split? If not, what standard should be adopted to create random splits and how many runs would be required?

\textcolor{nblue}{Answering these questions is critical to the progress of zero-shot learning, but especially in \gls{zsar} because processing videos is more time consuming and requires more hardware infrastructure than processing images. There is no discussion of acceptable classification accuracy, reaction time or resource efficiency in the literature.}

We conclude this section by pointing out an interesting and little explored problem that is recognizing whether an example is known or unknown and, based on this information, deciding which approach is more appropriate to try to recognize it. \textcolor{nblue}{Currently, we find only the works described by~\citet{roitberg:2018-ref56} and~\citet{mandal:2019-ref72} to consider both problems jointly}.

\section{Conclusions}
\label{sec:conclusion}

We presented a survey of available \gls{zsl} methods for action recognition in videos that describes several techniques used to perform visual and semantic extraction. We also presented several methods that employ these features and bridge the semantic gap. A comprehensive description of databases and their main applications is provided.

An analysis of the results was presented along with a discussion of the experimental protocols, from which we can highlight a number of conclusions. First, it is very difficult to compare experimental results since many of them use only one or two specific datasets (for instance, KTH, Weizmann, Charades, Breakfast, MPII Cooking Composites, UCF50) and do not follow the same protocol due to, for instance, differences in split sizes or random runs. To provide a fair comparison, we estimated the mean accuracy of each experiment using the available information and were able to compare experiments that reported standard deviation.

\textcolor{nblue}{The best results used combinations of features~\citep{wang:2017-ref18}, generative models~\citep{mandal:2019-ref72}, and action-object relationships~\citep{gao:2019-ref70,poppe:2010-ref62}. Multi-modal approaches (e.g., \citep{zhang:2018-ref28}) also presented promising results, although they are not comparable to most studies due to differences in the experimental protocol.}

When comparing the inductive against transductive setting, the results showed that the latter always presented better performance. Although they are not scalable, attributes showed superior results than word vectors, which demonstrates the need to extract high-level semantic information from videos. Finally, it is necessary to further investigate various protocol setups using state-of-the-art methods to identify the best configurations and the criteria for generating the splits, whether fixed or random.

\singlespacing

\section*{References}

\bibliographystyle{model2-names}
\bibliography{zsl}

\end{document}